\documentclass[pmlr,twocolumn,10pt]{jmlr}

\mlhtrack{proceedings}
\newif\iffinal
\finaltrue

\iffinal
  \ifmlhneedspmlr
    \jmlrvolume{346}
    \jmlryear{2026}
  \fi
  \jmlrworkshop{Machine Learning for Health (ML4H) 2026}
\else
  \jmlrproceedings{}{Submitted to ML4H 2026: \mlhtrackname}
  \jmlrworkshop{Machine Learning for Health (ML4H) 2026}
\fi

\usepackage{array}
\usepackage{booktabs}
\usepackage{microtype}
\usepackage{xurl}
\usepackage{cleveref}

\hypersetup{hidelinks}
\crefname{figure}{Fig.}{Figs.}
\Crefname{figure}{Fig.}{Figs.}

\DeclareUnicodeCharacter{2212}{\ensuremath{-}}
\newcommand{\yes}{\textbf{yes}}
\newcommand{\plaus}{plaus.}
\newcommand{\ourbench}{$\star$}
\newcommand{\CD}{\texttt{CD}}
\newcommand{\CDC}{\texttt{CD Complete}}
\newcommand{\CDI}{\texttt{CD Incomplete}}
\newcommand{\CDIU}{\texttt{CD Incomplete-Undeterminable}}
\newcommand{\CDID}{\texttt{CD Incomplete-Determinable}}
\newcommand{\MA}{\texttt{MA}}
\newcommand{\MAA}{\texttt{MA A}}
\newcommand{\MANA}{\texttt{MA NA}}
\newcommand{\MAAP}{\texttt{MA AP}}
\newcommand{\MANAP}{\texttt{MA NAP}}
\newcolumntype{L}[1]{>{\raggedright\arraybackslash}p{#1}}

\newif\ifdraftmarkup
\draftmarkupfalse

\title[Guardrail Erosion in Biomedical LLMs]{Guardrail Erosion in Domain Fine-Tuned Large Language Models: A Paired Audit of Knowledge-Centric Abstention in Biomedical LLMs}
\author{%
\Name{Stephen Peng}\nametag{\textsuperscript{1}\thanks{Equal contribution.}}\quad
\Name{Vineet Jagadeesan Nair}\nametag{\textsuperscript{1}\footnotemark[1]}\quad
\Name{Noah Broestl}\nametag{\textsuperscript{1,2}\footnotemark[1]}\\
\Name{Anne Friederike Kleppe}\nametag{\textsuperscript{1}}\quad
\Name{Steven Mills}\nametag{\textsuperscript{1}}\quad
\Name{Adi Zolotov}\nametag{\textsuperscript{1}}\\
\addr \textsuperscript{1}BCG X AI Science Institute, Boston Consulting Group\\
\addr \textsuperscript{2}Centre for Human-Inspired Artificial Intelligence, University of Cambridge\\
\normalfont\Email{\{Peng.Stephen,JagadeesanNair.Vineet,Broestl.Noah,Kleppe.Anne\}@bcg.com}\\
\Email{\{Mills.Steven,Zolotov.Adi\}@bcgfed.com}
}

\begin{document}

\maketitle

\begin{abstract}
The number of domain fine-tuned models, particularly in the biomedical domain, is rapidly expanding. Fine-tuning promises high-performing models at a fraction of the cost of training a base model, but it can also produce unintended behavioral changes. Multiple studies have shown that safety guardrails can erode with fine-tuning, yet large-scale studies across model architectures, data mixtures, and fine-tuning methodologies remain limited. Where prior work in this space has studied value-centric refusal, this work focuses on knowledge-centric abstention. We audit 25 biomedical domain fine-tuned large language models against 11 paired base checkpoints on MedAbstain and ClinDet-Bench to evaluate the impact of fine-tuning on knowledge-centric abstention.
We find that defined appropriate abstention measured by these benchmarks, already low in base models (median 8.3\% on MedAbstain), falls further after fine-tuning (median 4.2\%, decreasing in 21 of 25 pairs on MedAbstain), and that the magnitude and direction of change are heterogeneous across pairs.
These results motivate fine-tuning methods that improve domain-specific capability while preserving appropriate caution.
\end{abstract}

\begin{keywords}
fine-tuning, domain adaptation, biomedical large language models, abstention, model auditing, safety evaluation
\end{keywords}

\ifmlhneedsstatements
\paragraph*{Data and Code Availability}
We use the publicly available MedAbstain and ClinDet-Bench datasets cited in the Methods. An anonymized reproduction repository is available at \url{https://anonymous.4open.science/r/guardrail-erosion-paper-8D9F}.

\paragraph*{Institutional Review Board (IRB)}
This study evaluates public benchmark datasets and public model checkpoints and does not involve human participants or identifiable private data; IRB approval was not required.
\fi

\section{{Introduction and Related Work}}
\label{sec:introduction}
Although fine-tuning is less computationally demanding than training a new foundation model, modifying model weights can produce difficult-to-predict behavioral changes, including erosion of safety guardrails. \citet{qietal2024finetuning} found that safety alignment could be substantially compromised with as few as 10 adversarial fine-tuning examples and also observed smaller but non-trivial degradation after fine-tuning on benign, utility-oriented datasets. \citet{yangetal2024shadow} similarly shows that a small malicious fine-tuning set can broadly subvert refusal behavior. A safety-landscape analysis showed that fine-tuning can compromise alignment by moving a model away from a locally stable ``safety basin'' in parameter space \citep{pengetal2024navigating}. Broader audits find that fine-tuning risk depends on the dataset, model, and safety dimension under study \citep{vishnubhotlaetal2026safety}, and narrow task training can also induce misalignment outside the fine-tuning domain \citep{betleyetal2025emergent}. These studies primarily evaluate value-centric harmfulness or refusal rather than evidence-sensitive abstention or uncertainty-awareness.

Closest to our setting, AbstentionBench evaluates abstention on unanswerable questions across post-training stages and finds degradation in several reasoning adaptations \citep{kirichenko2026abstentionbench}. Our contribution is complementary, extending their analysis via a paired audit of adapted models in the biomedical domain, and evaluating caution jointly with capability, calibration, and commitment under cohort-level inference.

Knowledge-centric abstention asks whether available evidence is sufficient to support an answer. In clinical settings, a model should abstain or request more information when decisive evidence is absent, but answer when evidence is sufficient. We use \emph{guardrail erosion} in this study to mean a reduction after domain adaptation in this benchmark-defined appropriate caution. We distinguish three primary dimensions and one related probability-distribution property. \emph{Capability} measures correctness on conditions with a clinically valid answer, \emph{caution} measures whether the model answers when evidence is sufficient and abstains when it is not, and \emph{calibration} measures whether choice-normalized answer-token confidence agrees with empirical correctness. We additionally evaluate \emph{commitment}, the concentration of the choice-normalized probability distribution; commitment does not imply calibration, because a model can become sharper while becoming more confidently wrong. Clinical uncertainty quality varies across models, specialties, and question types \citep{testonicalixto2026mind}, motivating paired analyses of whether domain adaptation moves these dimensions together or separately.

\paragraph{{Contributions.}}
\begin{itemize}
    \item First, we conduct a paired audit of 25 biomedical domain fine-tuned models and the 11 base checkpoints from which they descend, spanning four model families and heterogeneous post-training recipes.
    \item Second, we jointly evaluate adaptation across capability, caution, calibration, and commitment, with question-level paired tests and cohort inference that accounts for shared base checkpoints.
    \item Third, we find that appropriate caution declines robustly on MedAbstain without a commensurate improvement in informedness, newly answered questions occupy a comparatively error-prone region, and answer-token commitment increases, including on shared errors.
    \item Fourth, and central to our findings, these effects are heterogeneous across model pairs, underscoring that biomedical fine-tuning should be evaluated as a multidimensional behavioral intervention rather than a single safety property.
\end{itemize}

\section{Methods}

\subsection{{Benchmarks, Model Pairs, and Protocol}}
\label{sec:exp-methods}
{MedAbstain (\MA) derives four variants from each medical multiple-choice item by combining original versus evidence-ablated questions with the absence versus presence of an abstention option: No-Abstention (\MANA), Abstention (\MAA), No-Abstention-Perturbed (\MANAP), and Abstention-Perturbed (\MAAP) \citep{machchaetal2026knowinga}. We use the public MedQA-USMLE half, containing 1,007 items per variant. ClinDet-Bench (\CD) uses optionally deterministic clinical scoring rules with optionally limited information and asks whether criteria are ``Met,'' ``Not Met,'' or ``Unable to determine'' \citep{watanabeetal2026clindetbencha}. We evaluated the 94 scenarios in public repository commit \href{https://github.com/yusukewatanabe1208/ClinDet_Benchmark/tree/1a05c632512be8270be863ed0d304c4742a535bd}{\texttt{1a05c63}} consisting of 32 Complete, 30 Incomplete-Determinable, and 32 Incomplete-Undeterminable scenarios. This pinned public snapshot contains 94 scenarios (while the ClinDet-Bench paper reports 188 evaluated cases)---our evaluation targets the smaller public release. \MA\ provides scale but imperfect perturbation validity, whereas \CD\ provides deterministic labels but limited sample size; see App. Table~\ref{tab:benchmarks} for the broader benchmark landscape.}

We evaluate 25 biomedical domain fine-tuned models and the 11 base models they were trained on, spanning Llama 2, Llama 3, Llama 3.1, and Gemma-3 (base-model repositories, model pairings, training recipes, and sources in App. Tables~\ref{tab:models}--\ref{tab:finetuned-models}). We select clinically fine-tuned models based on three practical criteria -- availability (model checkpoints available online), documented base-model lineage (explicitly disclosed except where marked as inferred in App. Table~\ref{tab:finetuned-models}), and at least partial disclosure of fine-tuning recipe -- before prioritizing models to ensure coverage across parameter scales and model families. All models were evaluated using plain-text completion prompts rather than model-specific chat templates and ended with an \texttt{Answer:} heading, reducing variation between models. We use greedy decoding with temperature 0 and a limit of 2,048 output tokens. Multiple-choice answers were parsed from the completion using waterfall logic from the most to least definitive answer output format, while unparseable or conflicting outputs were scored as incorrect and non-abstaining. Appendix~\ref{app:lineage-reproducibility} records the exact prompt template and parser used. Appendix~\ref{app:native-template} reports a sensitivity analysis in which the nine pairs with a chat-capable base checkpoint were re-evaluated under their native chat templates; the direction of the primary abstention effect remains the same, and no outcome shows significant changes under the native template.

The recipe families studied include: continued pre-training (CPT), supervised fine-tuning (SFT), reinforcement learning (RL), direct preference optimization (DPO), parameter-efficient fine-tuning (PEFT), and model merging (MERGE). For probability-based metrics, we identify the first generated answer token and sum the probabilities of token variants representing each answer choice within the saved top-20 token list. We then rescale these choice probabilities to sum to one. These metrics therefore measure normalized confidence relative to the available answer choices rather than the model's full output distribution, by excluding probability assigned to other tokens (Appendix~\ref{app:lineage-reproducibility}). The evaluation consumed approximately 38 L40S and 28 H100 GPU-hours using vLLM (hardware details appear in Appendix~\ref{app:lineage-reproducibility}).

\subsection{Evaluation Metrics}

For metric $m$ and benchmark condition $v \in \{\mathrm{MA}, \mathrm{CD}\}$, we compute the absolute change within each matched model pair:
\begin{equation}
\label{eq:delta_general}
    \Delta m_v=m_{f,v}-m_{b,v},
\end{equation}
where $b$ and $f$ denote the base and fine-tuned roles. Proportion changes are reported in percentage points (pp); relative changes are supplementary and calculated within each pair.

Capability is measured using accuracy on conditions with a clinically valid gold answer. Let $U_{r,v}$ denote the abstention rate for role $r$ in condition $v$; for \CD, selecting ``Unable to determine'' constitutes abstention. Appropriate caution is measured on \MAAP\ and \CDIU\ . Abstention on \MAA\ or choosing ``Unable to determine'' on \CDID\ instead measure excessive caution. To measure whether a model distinguishes cases in which abstention is and is not appropriate, we adapt Youden's $J$ statistic \citep{youden1950index}:
\begin{equation}
\label{eq:informedness_j}
    J_r^{\mathrm{MA}}=U_{r,\mathrm{AP}}-U_{r,\mathrm{A}},
    \qquad
    J_r^{\mathrm{CD}}=U_{r,\mathrm{Undet}}-U_{r,\mathrm{Det}}.
\end{equation}
Larger $J$ indicates greater informedness.

We also partition \MAA\ questions into always answered, newly answered, newly abstained, and both abstained according to the base and fine-tuned models' abstention decisions. The newly answered accuracy gap compares the fine-tuned model's accuracy on newly answered questions with its accuracy on always-answered questions. The symmetric newly abstained gap compares the base model's accuracy on newly abstained questions with its accuracy on always-answered questions. We transfer these partitions to \MANA, where both models must answer, to evaluate capability on identical underlying questions.

Calibration and commitment are evaluated on \MANA\ and \CDC\ using expected calibration error over ten equal-width confidence bins (ECE10), mean maximum-choice probability, normalized entropy, and the Area Under the Receiver Operating Characteristic (AUROC) curve of confidence for discriminating correct from incorrect predictions. We also measure confidence on questions both models answer incorrectly. These metrics use only aligned questions with valid choice probabilities and gold labels. Complete definitions, responsiveness diagnostics, and secondary outcomes appear in Appendix~\ref{app:metric-definitions}.

\subsection{Statistical Inference}

We test five hypotheses:
\begin{itemize}
    \item Whether fine-tuning reduces appropriate caution (\textbf{H1})
    \item Whether fine-tuning changes informedness and, in a supplementary MedAbstain analysis, whether informedness improves among pairs that lose appropriate caution (\textbf{H2})
    \item Whether newly answered questions have lower observed accuracy than questions answered by both roles, with a symmetric analysis of newly abstained questions (\textbf{H3})
    \item Whether answer-token commitment and calibration change and whether confidence-based error discrimination changes (\textbf{H4})
    \item Whether capability and caution changes covary and whether caution effects are heterogeneous across model pairs (\textbf{H5}).
\end{itemize}

We distinguish two units of inference. Within a fine-tune/base pair, the benchmark item is the unit of analysis. We use exact McNemar tests for aligned binary outcomes, Fisher exact tests for correct/incorrect comparisons between disjoint coverage-transition partitions, and paired item bootstraps for continuous and cross-condition quantities. Across the model cohort, the fine-tune/base pair is the unit of inference. For marginal change outcomes, each pair contributes one $\Delta m_v$, and we use two-sided Wilcoxon signed-rank tests to assess whether the distribution of pair-level effects is symmetric about zero. Capability--caution associations and moderator analyses use permutation-based correlations or group contrasts, while Cochran's $Q$ assesses effect heterogeneity. We do not pool benchmark items across model pairs for inferential claims.

We use Benjamini--Hochberg (BH) correction within prespecified test families to account for multiple hypothesis testing. We report unadjusted p-values as $p$ and BH-adjusted values as $q_{\mathrm{BH}}$, with $q_{\mathrm{BH}}<0.05$ considered statistically significant.
For the main results, BH correction is applied across 15 prespecified tests in the appropriate-caution erosion (\textbf{H1}), informedness-change (\textbf{H2}), coverage-change accuracy (\textbf{H3}), commitment and calibration (\textbf{H4}), and capability--caution coupling and heterogeneity (\textbf{H5}) hypotheses.
We also report a secondary commitment and calibration (\textbf{H4}) correction across ten probability-based outcomes: ECE10, normalized entropy, maximum-choice confidence, shared-wrong confidence, and confidence AUROC on \MANA\ and \CDC.
Because the five \MANA\ outcomes appear in both sets, primary conclusions use their adjusted values from the 15-test family.
Within-pair and exploratory moderator analyses are corrected separately within their respective test families.

Because the 25 fine-tuned models descend from only 11 base checkpoints, model pairs sharing a base are not fully independent. For Wilcoxon-based cohort outcomes, we therefore also use a two-stage hierarchical bootstrap that resamples base checkpoints and then fine-tuned descendants within each selected base. We denote its two-sided base-clustered percentile-sign p-value by $p_{\mathrm{cluster}}$. A cohort-level change is treated as robust when its effect direction is consistent, $q_{\mathrm{BH}}<0.05$ in the governing family, and $p_{\mathrm{cluster}}<0.05$. The \textbf{H5} association and heterogeneity analyses use their corresponding permutation tests and Cochran-$Q$ procedures and are reported separately from this clustered robustness criterion.

Appendix~\ref{app:statistical-methods} describes the confirmed-overlap-excluded, thin-cell-excluded, and relative-change sensitivity analyses. It also gives the complete tests, correction families, clustering procedure, heterogeneity weights, and exploratory moderator analyses of model size, base alignment, training strength, and fine-tuning recipe. Because this is an observational audit of publicly released models ``in the wild'', these moderator characteristics are confounded.
\section{{Results}}
\label{sec:results}

Unless otherwise stated, reported effect sizes are medians across fine-tune/base pairs, changes are fine-tuned minus base, and proportion changes are reported in percentage points (pp). We distinguish findings that survive both false-discovery-rate BH correction and base-clustered inference from directional results that attenuate under robustness analyses.

\subsection{Knowledge-Centric Caution Erodes after Fine-Tuning (H1)}
\begin{figure*}[htbp]
    \centering
    \includegraphics[width=0.8\textwidth,trim=0bp 303bp 0bp 0bp,clip]{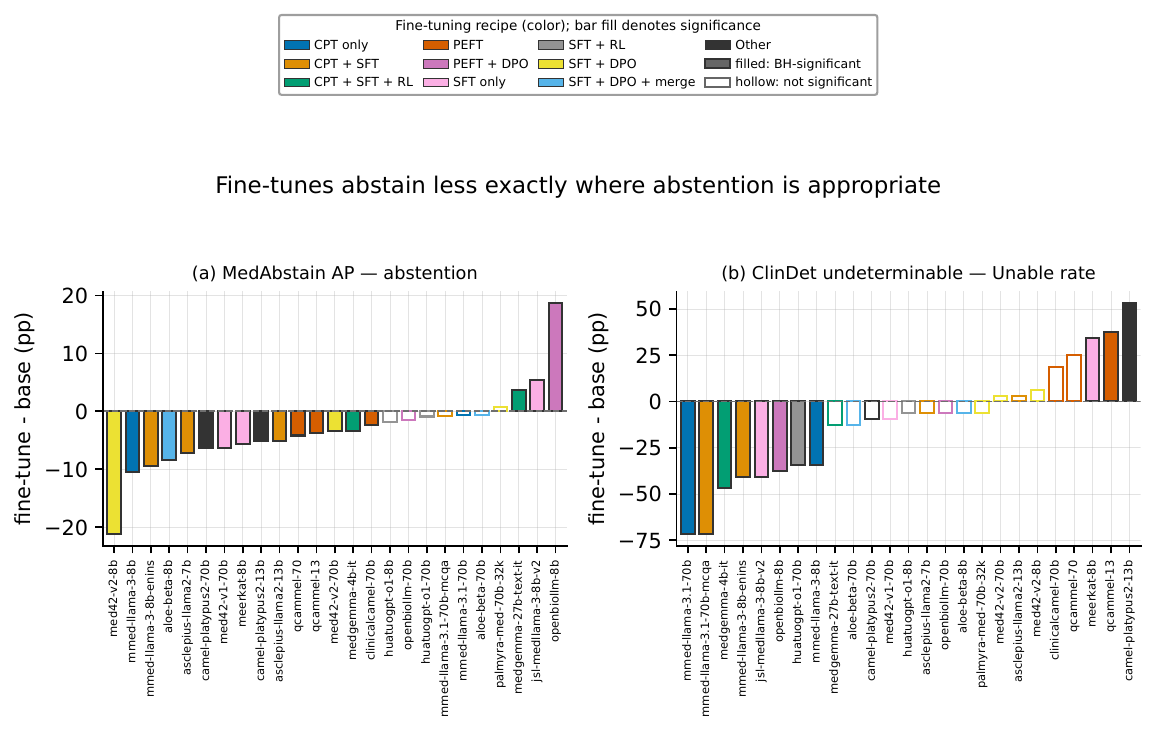}\\
    \includegraphics[width=0.8\textwidth,trim=0bp 4.957bp 0bp 100bp,clip]{figures/fig01_abstention_change_absolute_paper.pdf}
    \caption{Fine-tuning generally reduces benchmark-defined appropriate abstention. Pair-level absolute change (fine-tune minus base, percentage points) in \MAAP\ abstention (left) and \CDIU\ ``Unable'' rate (right), one bar per model pair. Filled bars denote BH-significant within-pair McNemar tests ($q_{\mathrm{BH}}<0.05$); hollow bars denote non-significant tests. The shared legend identifies fine-tuning recipe families.}
    \label{fig:pairlevel_abstention_change}
\end{figure*}

After fine-tuning, models were less likely to abstain appropriately on ambiguous questions. On \MAAP, across 25 pairs, the median change in \MAAP\ abstention was $-3.48$ pp; appropriate abstention decreased in 21 pairs and increased in four (\cref{fig:pairlevel_abstention_change}).
This shift remained significant after correction and base-clustered inference ($p=0.0038$, $q_{\mathrm{BH}}=0.009$, $p_{\mathrm{cluster}}=0.003$; base-clustered 95\% CI: $-5.66$ to $-0.79$ pp). Median base and fine-tuned levels were 8.3\% and 4.2\%, and the median within-pair relative decrease was 45.0\% (App. Fig.~\ref{fig:abstention_levels}). The result is insensitive to parse failures (App.~\ref{app:protocol-sensitivity}), and residual answerability of some \MAAP\ items is unlikely to explain the effect (Sec.~\ref{sec:limitations}; App. Fig.~\ref{fig:ap_answerable_rebuttal}). \CD\ was directionally similar but inconclusive. Across 25 pairs, the median change in \CDIU\ ``Unable to determine'' rate was $-6.25$ pp (\cref{fig:pairlevel_abstention_change}), but the effect did not survive correction or base-clustered inference ($p=0.061$, $q_{\mathrm{BH}}=0.083$, $p_{\mathrm{cluster}}=0.182$). The robust erosion claim is therefore specific to \MA; \CD\ provides directional, albeit inconclusive, corroboration.

\subsection{{Reduced Caution Is Not Explained by Greater Informedness (H2)}}
\label{sec:explain-caution}

\begin{figure*}[htbp]
    \centering
    \includegraphics[width=0.8\textwidth,trim=0bp 4.105bp 0bp 30.5bp,clip]{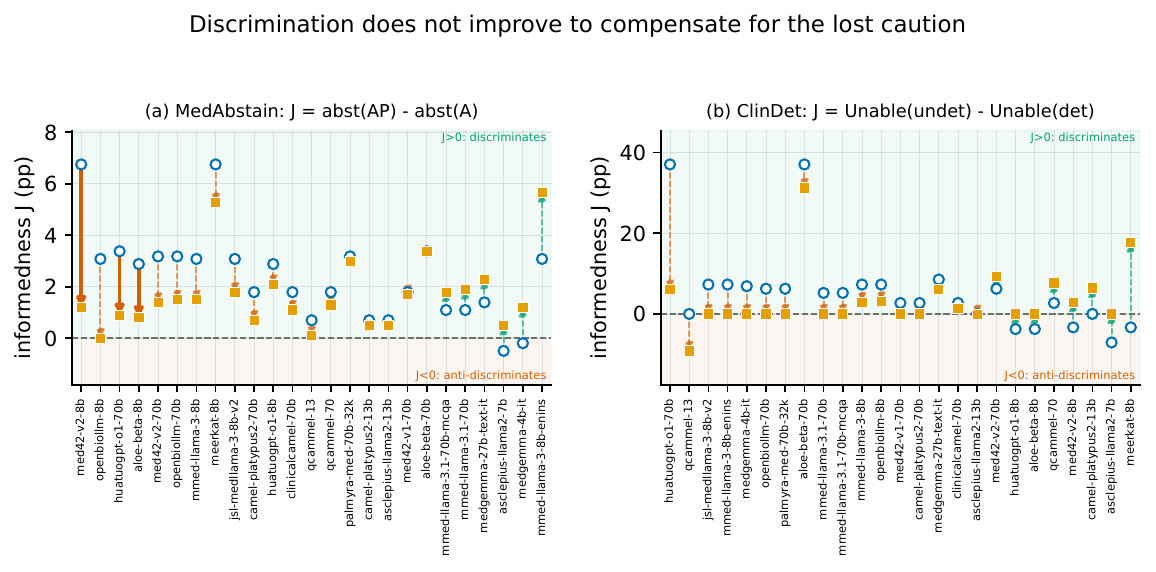}
    \caption{{Discrimination does not improve to compensate for the lost caution. Informedness $J$ by model, for \MA\ ($J^{\mathrm{MA}}=U_{\mathrm{AP}}-U_{\mathrm{A}}$, left) and \CD\ ($J^{\mathrm{CD}}=U_{\mathrm{Undet}}-U_{\mathrm{Det}}$, right). $J>0$ indicates that a model abstains more often when appropriate than when it is not. Open circles mark base checkpoints and filled squares mark adapted checkpoints; each vertical arrow connects one pair. Solid, heavier arrows indicate pair-level BH significance, while dashed, lighter arrows indicate non-significance.}}
    \label{fig:informedness_j}
\end{figure*}

A reduction in abstention could be beneficial if fine-tuning improved the distinction between answerable and abstention-appropriate questions. However, we found little evidence for this explanation (\cref{fig:informedness_j}).
In fact, we observed a potential decrease in the model's ability to discriminate between these questions after fine-tuning. \MA\ informedness decreased by a median of $\Delta J^{\mathrm{MA}}=-0.60$ pp, but the change was nominally significant only before correction and did not survive it ($p=0.034$, $q_{\mathrm{BH}}=0.052$) or base-clustered inference ($p_{\mathrm{cluster}}=0.100$). \CD\ informedness also showed no reliable change (median $\Delta J^{\mathrm{CD}}=-2.71$ pp; $p=0.158$, $p_{\mathrm{cluster}}=0.195$). Weak responsiveness to evidence removal was already present in base models (App. Fig.~\ref{fig:responsiveness_targeting}). We therefore interpret weak informedness as an inherited limitation that fine-tuning does not reliably correct, not as a mechanism whose further deterioration is established across model families.

\subsection{{Fine-Tuned Models Perform Worse on Newly Answered Questions (H3)}}
\label{sec:boundary}

\begin{figure*}[htbp]
    \centering
    \includegraphics[width=0.8\textwidth,trim=0bp 4.041bp 0bp 29.5bp,clip]{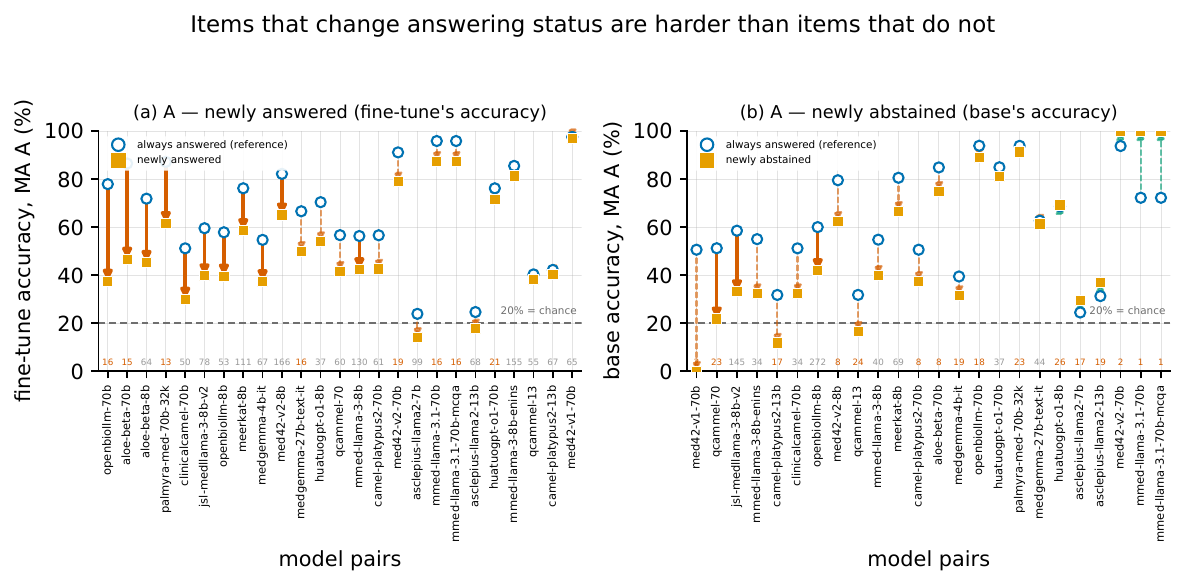}
    \caption{{Items that change answering status are harder than items that do not. Accuracy on \MAA\ questions newly answered by the fine-tuned model (left, fine-tune accuracy) versus accuracy on the always-answered reference partition, and the base model's accuracy on questions newly abstained by the adapted model (right) versus the same reference partition. Adjacent counts give partition size per pair; the dashed line marks chance accuracy (20\%, among five clinical options conditional on answering). Solid lines connect pairs with a BH-significant change.}}
    \label{fig:ma_partition_accuracy}
\end{figure*}

Questions brought into coverage by fine-tuning were relatively more difficult for the model compared to questions that were always answered. On \MAA, the fine-tuned model's accuracy on newly answered questions was a median of 14.98 pp lower than its accuracy on questions answered by both models, a relative deficit of 24.8\% (\cref{fig:ma_partition_accuracy}).
This was the largest and most statistically robust effect in our primary hypothesis family. We note that since the partition is conditioned on the base model's abstention decisions, newly answered questions are expected to be harder than always-answered ones; however, we can interpret the effect as identifying the model-adjudicated difficulty of questions on which it loses caution. The corresponding statistics were $p<0.0001$, $q_{\mathrm{BH}}=0.00009$, $p_{\mathrm{cluster}}<0.001$, with a base-clustered 95\% CI of $-17.91$ to $-8.36$ pp. The effect persisted in analyses excluding confirmed overlaps and thin cells (App.~\ref{app:robustness-sensitivity}).

The symmetric transition showed that the base model's accuracy on newly abstained questions was 11.47 pp lower than its accuracy on always-answered questions ($p=0.014$, $q_{\mathrm{BH}}=0.023$), though the result narrowly misses the base-clustered criterion ($p_{\mathrm{cluster}}=0.051$; \cref{fig:ma_partition_accuracy}). The two negative gaps describe different behaviors near the competence boundary: newly answered questions represent a loss of caution in a comparatively error-prone region, whereas newly abstained questions represent added caution on questions that the base model answered relatively poorly. On \MANA, where the model cannot abstain, no fine-tuned-versus-base accuracy change within these partitions survived correction (App. Fig.~\ref{fig:na_partition_accuracy}). Therefore we can only conclude that coverage-changing items are comparatively hard, not that fine-tuning uniformly reduced knowledge on those items.

\subsection{Fine-Tuning Increases Commitment and Worsens Calibration on MedAbstain (H4)}
\label{sec:commitment}

\begin{figure*}[htbp]
    \centering
    \includegraphics[width=0.95\textwidth,trim=0bp 19.448bp 0bp 29.5bp,clip]{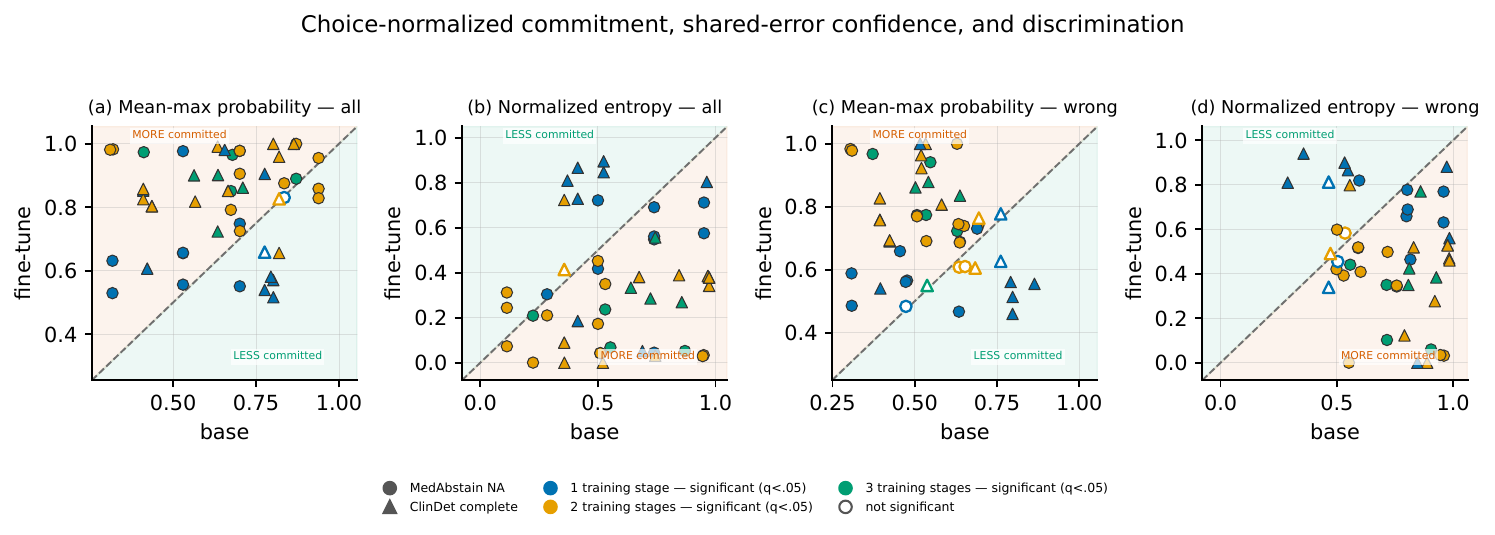}
    \caption{{Choice-normalized commitment and shared-error confidence. Fine-tune versus base mean maximum-choice probability and normalized entropy, each computed over all answers and then restricted to questions both models answer incorrectly (left to right), on \MANA\ and \CDC. Restricting to shared errors shows that any sharpening is not explained by the fine-tuned model simply being correct more often. Marker shape denotes the benchmark, color denotes the number of fine-tuning stages, and fill denotes whether the pair was BH significant.}}
    \label{fig:confidence_grid}
\end{figure*}

Overall, models exhibited worse calibration and greater commitment to answers after fine-tuning. On \MANA, normalized entropy decreased by 0.191 while maximum-choice probability increased by 0.127 (\cref{fig:confidence_grid}); both changes survived correction and base-clustered inference (entropy: $q_{\mathrm{BH}}<0.001$, $p_{\mathrm{cluster}}=0.001$; probability: $q_{\mathrm{BH}}<0.001$, $p_{\mathrm{cluster}}=0.002$). Shared-wrong confidence increased by 0.112 ($q_{\mathrm{BH}}<0.001$, $p_{\mathrm{cluster}}=0.001$), showing greater commitment on the same incorrect items rather than a shift toward easier questions.

\MANA\ ECE10 increased by 0.040 and survived both the 15-claim correction and base-clustered inference ($q_{\mathrm{BH}}=0.009$, $p_{\mathrm{cluster}}=0.040$). Confidence AUROC did not change reliably.
On \CDC, mean answer-token confidence increased by 0.185 ($q_{\mathrm{BH}}=0.018$), normalized entropy decreased by 0.307 ($q_{\mathrm{BH}}=0.019$), shared-wrong confidence increased by 0.225 across 23 estimable pairs ($q_{\mathrm{BH}}=0.019$), and AUROC increased by 0.089 ($q_{\mathrm{BH}}=0.060$). Only the confidence increase met the prespecified base-clustered robustness criterion ($p_{\mathrm{cluster}}=0.035$, $0.089$, $0.094$, and $0.063$, respectively), and \CDC\ ECE10 did not change reliably.
Corresponding ECE10 comparisons for other \MA\ variants appear in App. Fig.~\ref{fig:ece_identity_appendix}.
Fine-tuning therefore changed the scale and concentration of choice-normalized answer-token probabilities more reliably than their ability to discriminate errors.

\subsection{{Capability and Caution Changes Are Heterogeneous (H5)}}

\begin{figure*}[htbp]
    \centering
    \includegraphics[width=0.9\textwidth,trim=0bp 7.16bp 0bp 18bp,clip]{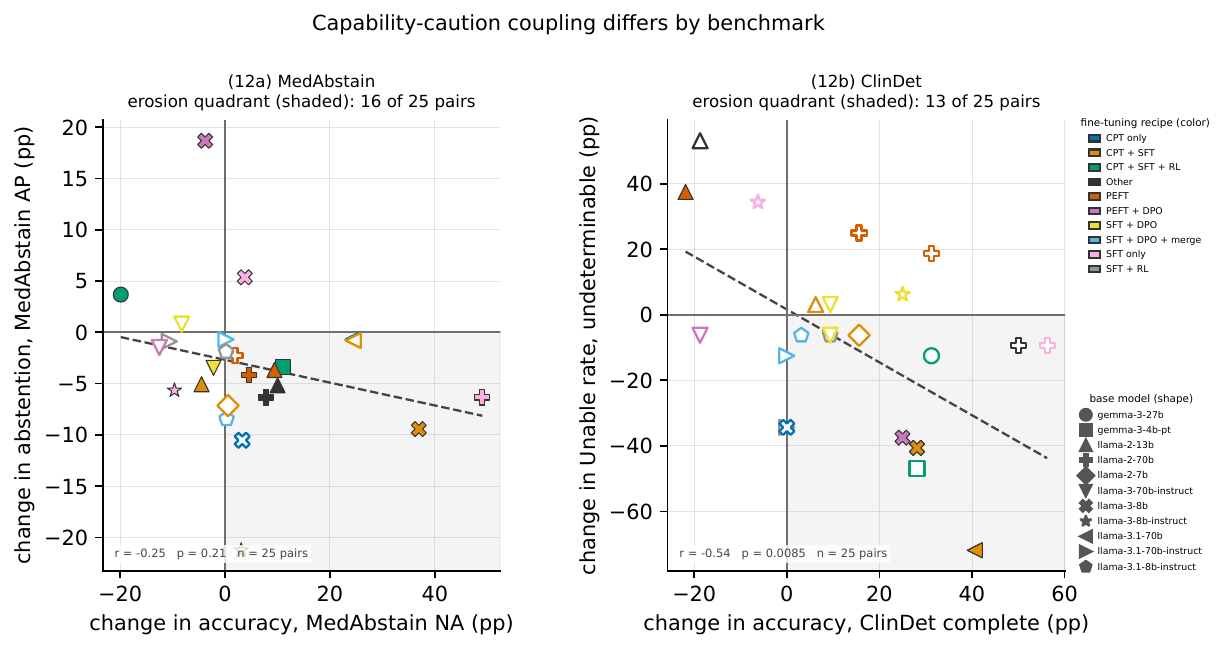}
    \caption{Capability--caution coupling differs by benchmark. Pair-level change in appropriate abstention plotted against pair-level change in capability, for MedAbstain (left: \MAAP\ abstention change vs.\ \MANA\ accuracy change, $r_{\mathrm{P}}=-0.25$) and ClinDet (right: \CDIU\ Unable-rate change vs.\ \CDC\ accuracy change, $r_{\mathrm{P}}=-0.54$). The shaded quadrant marks the erosion pattern (more accurate and less cautious); points are colored by fine-tuning recipe and shaped by base model.}
    \label{fig:capability_caution_scatter}
\end{figure*}

Capability and caution changes were not uniformly coupled. On \MA, their correlation was weak and non-significant ($r_{\mathrm{P}}=-0.249$, permutation $p=0.210$, $q_{\mathrm{BH}}=0.225$). Sixteen pairs increased capability while decreasing caution; one increased both; three decreased capability while increasing caution; and five decreased on both dimensions (\cref{fig:capability_caution_scatter}). \CD\ showed a significant negative association ($r_{\mathrm{P}}=-0.538$, permutation $p=0.008$, $q_{\mathrm{BH}}=0.028$).

Both benchmarks showed substantial between-pair heterogeneity ($I^2=96.3\%$ for \MA\ and $93.4\%$ for \CD; complete $Q$ statistics and weighting details are in App.~\ref{app:results-hetero}). These results do not establish a universal trade-off where capability gains require less caution. Instead, adaptation effects remain idiosyncratic to different configurations of the base and fine-tuned model's training processes and the two dimensions move separately. None of the exploratory moderators in our analyses -- recipe family, base-model alignment, training strength, or parameter count -- was significant after multiplicity correction. Small subgroup sizes and confounding among model family, training data, and optimization choices limit attribution (App. Figs.~\ref{fig:moderators_recipe}--\ref{fig:recipe_accuracy_caution}).

\section{Discussion}

The primary takeaway from our analysis is not that every biomedical fine-tune becomes less cautious.
Four \MA\ pairs showed increased appropriate caution, the capability--caution association differed by benchmark, and no moderator survived correction.
Rather, we observe that domain adaptation can move correctness, answer propensity, and probability commitment without a uniform relationship across benchmarks.
We observe that when models become less cautious, changes in abstention tend to occur near the capability boundary, on questions that are comparatively difficult for the model to answer. Reporting accuracy alone can therefore hide a shift toward undesirable behavior such as answering more difficult cases with greater commitment.

The observational data are insufficient to establish mechanisms but suggest three hypotheses for further study. First, answer-always supervision, such as exam-style question answering, may reinforce answering (especially when the baseline does not possess strong caution guardrails). Second, refusal-reduction objectives may suppress knowledge-centric abstention as well as unwanted domain refusal, although our available evidence is limited to two models from one release family \citep{christopheetal2024med42v2}. Arditi et al. identify a direction associated with value-centric refusal rather than knowledge-centric abstention, so their mechanism is only a possible conceptual bridge but does not establish this hypothesis \citep{arditietal2024refusal}. Finally, post-training may enlarge answer-token confidence margins without improving error discrimination, consistent with increased commitment on shared errors. Controlled fine-tuning that holds the base model fixed while varying data composition and preference objectives is required to distinguish these explanations, and we aim to explore this in future work.

For model development, the practical implication is to evaluate base and adapted checkpoints as matched pairs and to report all four dimensions explored here. Capability should be evaluated on answerable conditions; caution should distinguish appropriate from excessive abstention; calibration should assess whether confidence reflects correctness; and commitment should measure whether predictions become sharper, including on shared errors. These checks do not guarantee clinical safety, but they expose behavioral changes that evaluating for accuracy alone would not detect.

For practitioners, we order our recommendations from most to least confident. (1)~Evaluate base and adapted checkpoints across multiple dimensions and as matched pairs; accuracy-only and single-model evaluations would have hidden every effect observed in this work. (2)~Measure the change in abstention on evidence-ablated conditions before releasing adapted models in environments where abstention is critical; appropriate abstention fell in 21 of 25 released pairs. (3)~Inspect the questions the adapted model newly answers to confirm capabilities remain within the acceptable range; in our study, these were a median 14.98~pp less accurate than always-answered questions. (4)~Track answer-token commitment on errors shared between the base and adapted models; sharpening without improved error discrimination indicates growing miscalibration. (5)~Do not assume any recipe family preserves appropriate abstention and calibration; no moderator survived correction in our analysis, warranting a case-by-case audit for each release.

\section{Limitations and Future Work}
\label{sec:limitations}

{The benchmarks analyzed in this work are limited in scope. \MA\ uses multiple-choice questions and LLM-generated evidence ablations rather than interactive clinical decisions. In the benchmark's human validation round, abstention was judged appropriate in 38 of only 49 labeled perturbed cases, so \MAAP\ does not fully reflect true clinical indeterminacy. Furthermore, many of the models we evaluated contain or presumably contain MedQA in their training or fine-tuning data, raising a risk of contamination when evaluating on a benchmark derived from MedQA. %
Contamination can affect answer propensity and confidence as well as correctness, and limitations in public disclosures mean a sensitivity fully free of contamination is impossible. We note that contamination and benchmark saturation are pervasive issues in this field, as open-source evaluation sets are typically adopted as training data for subsequent models. In future work, we plan to develop a contamination-free benchmark to investigate these results further. \CD\ provides deterministic labels but contains only 94 cases (and only 32 \CDIU\ cases), limiting the ability to draw significant conclusions. Neither benchmark establishes how a model would defer, request information, or escalate in a deployed workflow, and our use of an explicit abstention option may not generalize to open-ended generation.}

Additionally, we note our experimental choices do not cover the full range of settings in which we may wish to evaluate these models.
A common completion prompt and greedy decoding improve comparability across pairs. To investigate the effect of the chat template, we conduct a native template sensitivity analysis on the nine model pairs released with chat templates; this analysis reproduces the direction of the primary effect (App.~\ref{app:native-template}), but decoding sensitivity and publisher system prompts remain untested.
These metrics use choice-normalized probabilities reconstructed from the saved top-20 tokens, which discards the probability lying outside the declared choice set. An exact full-vocabulary re-scoring of ten pairs (App.~\ref{app:full-vocab-rescoring}) leaves the direction of the effect unchanged in every pair, preserves their ordering, and shows that the mass outside the choice set does not shift systematically with adaptation; renormalization does, however, inflate the shared-error commitment change by a median of 0.7 points. We therefore describe these quantities as choice-normalized answer-token confidence, and note that this smaller analysis is not powered to retest the effect itself.
These constraints support a paired benchmark audit and not claims of clinical deployment readiness. Finally, we note the model audit is observational.
Model family, base alignment, parameter count, training data, and optimization recipe are confounded; public disclosures are incomplete; and possible benchmark contamination cannot be eliminated.
We earmark for future work a controlled empirical fine-tuning study to perform a more causal analysis of the drivers of erosion discussed here.

\section{Conclusion}

Across 25 paired biomedical adaptations, fine-tuning robustly reduced benchmark-defined appropriate caution on MedAbstain, while ClinDet-Bench was directionally consistent but inconclusive. Newly answered questions were comparatively error-prone, and choice-normalized answer-token commitment increased even on shared errors without a robust improvement in error discrimination. Because capability, caution, and commitment changed heterogeneously, biomedical fine-tuning should be evaluated as a multidimensional behavioral intervention. Controlled studies are now needed to identify which data and objectives preserve appropriate caution while improving capability.

\paragraph*{LLM Usage Statement}
We used LLMs for editing, LaTeX formatting, and coding assistance. We reviewed and validated all generated content. The authors designed, executed, analyzed, and verified the study.

\bibliography{references_zotero}

\clearpage
\appendix
\section{{Additional Analyses}}

\subsection{{Additional Background}}

The proliferation of domain-specialized foundation models, particularly in biomedical science and pharmaceutical R\&D, has been rapid \citep{delileetal2025foundation,2026generative}. Because training foundation models from scratch is costly, fine-tuning and parameter-efficient adaptation provide practical mechanisms for specializing shared base models while updating far fewer parameters and reducing memory requirements \citep{bommasanietal2022opportunities,huetal2021loraa}. Biomedical adaptations now span medical question answering and clinical reasoning, electronic health record analysis, drug discovery, and pharmaceutical R\&D workflows \citep{sellergrenetal2025medgemma,delileetal2025foundation,2026generative}. The result is a heterogeneous ecosystem of systems built atop a small number of shared foundation models, with widely varying training recipes and methodologies.

Fine-tuning adapts a general-purpose language model by updating its parameters on domain- or task-specific data. Supervised fine-tuning (SFT) learns from input--output demonstrations, while continued pre-training (CPT) extends next-token training on an additional unlabeled corpus; biomedical models commonly combine these stages to inject domain knowledge and then shape instruction-following behavior \citep{ouyangetal2022training,qiuetal2024building}. Parameter-efficient methods reduce the number of updated weights. Low-Rank Adaptation (LoRA) freezes the backbone and learns low-rank update matrices \citep{huetal2021loraa}; quantized LoRA applies the same principle to a quantized backbone \citep{dettmersetal2023qlora} and is used by several clinical models in our cohort \citep{tomaetal2023clinical}.

Preference optimization is a distinct post-training stage. Reinforcement learning from human feedback (RLHF) typically fits a preference or reward model and then optimizes the policy, as in the PPO-based InstructGPT pipeline \citep{ouyangetal2022training}. Direct Preference Optimization (DPO) instead learns directly from preference pairs without fitting an explicit reward model or running reinforcement learning \citep{rafailovetal2023direct}. These stages can be composed with CPT and SFT, and SFT targets may include chain-of-thought rationales intended to strengthen task-specific reasoning \citep{garcia-gasullaetal2026aloe,chenetal2024huatuogpto1}. Because each stage changes model behavior under a different training signal, post-training ``recipe'' denotes both the sequence of stages and the data used within them.

\subsection{Prompts and Parsers}
\label{app:lineage-reproducibility}

\paragraph{Exact completion prompts.} MedAbstain used the following template, with one choice line per labeled option:
\begin{quote}
\small\ttfamily
Question: $\langle$question$\rangle$\\
Choices:\\
$\langle$letter$\rangle$. $\langle$choice text$\rangle$\\
Answer:
\end{quote}
ClinDet used:
\begin{quote}
\small\ttfamily
Based on the following case description, determine whether the patient meets the criterion '$\langle$criterion$\rangle$' according to $\langle$score name$\rangle$. Answer with your final judgment as 'Met', 'Not met', or 'Unable to determine'.\\[2pt]
Case: $\langle$case presentation$\rangle$\\[2pt]
Answer:
\end{quote}

\paragraph{Answer parser.} The MedAbstain waterfall checks, in order, a leading choice letter, an explicit ``answer is/answer:'' letter, the first standalone A--F letter, and the earliest sufficiently long exact choice-text mention. The ClinDet strict parser first reconciles a leading label with the last explicit final-answer label, marks disagreement as a conflict, and otherwise falls back in order to ``Unable to determine,'' ``Not met,'' and ``Met'' phrase patterns. If no valid label remains, the output is scored as incorrect and non-abstaining.

\paragraph{Choice-probability extraction.} At each generated position, we retained log probabilities for the top 20 tokens. To identify an answer position, we reconstructed the greedy token sequence from the highest-log-probability token at each position and verified that it matched the saved completion prefix. We then used the first position at which the generated token was an answer letter or the first token of a \CD\ label (``Met,'' ``Not,'' or ``Unable''/``Un''). For each choice $k$, we summed the probability over its matching case- and whitespace-normalized token forms in the top 20 tokens and renormalized these masses across the choice options; choices absent from the top-20 set received zero mass. Thus, choice probabilities are restricted and renormalized over answer choices at the first relevant output position, not full-sequence probabilities.

\paragraph{Compute.} Small models ($\leq 13$B parameters) were run on single-GPU (NVIDIA L40S) instances, while the larger models ($27$B-$70$B parameters) were run on eight-GPU (8x H100 80GB) instances, using vLLM 0.26.0 and 0.27.1 respectively.

\subsubsection{Parse-Failure Sensitivity}
\label{app:protocol-sensitivity}

On \MAAP, median parse-failure rates were 0\% for both roles, while mean rates were 0.04\% for base and 2.10\% for adapted checkpoints; Asclepius-Llama2-13B, Asclepius-Llama2-7B, and MedGemma-4B-IT had adapted-checkpoint \MAAP\ parse-failure rates of 29.1\%, 21.4\%, and 1.6\%. Excluding parse-failed items within each role left the cohort result essentially unchanged (median $\Delta$ abstention $-3.48$ pp; Wilcoxon $p=0.0038$; base-clustered percentile-sign $p=0.003$; clustered 95\% CI: $-5.66$ to $-0.79$ pp), and so we conclude parse failures do not have a meaningful effect on our abstention calculations.

\subsubsection{Native Chat-Template Sensitivity}
\label{app:native-template}
The primary protocol evaluates every checkpoint with the same plain-text completion prompt, which controls prompt format within each pair but raises the question of whether the result survives evaluation under each checkpoint's own chat template. We therefore repeated the analysis on the subset of pairs for which both the base and the adapted model can be evaluated under its native chat template.

\paragraph{Native-template subset.} We perform the analysis on checkpoints which release a documented chat template on both sides of the base-adapted pair. Sixteen of the 25 primary pairs are derived from pretrained-only base checkpoints (Llama-1, Llama-2, Llama-3-8B, Llama-3.1-70B, Mistral-7B, and Gemma-3-27B-PT) that publish no chat template; evaluating only the adapted side with a native template would confound differences with within-pair differences in prompt templates. The remaining nine pairs span four instruction-tuned base checkpoints: Llama-3-8B-Instruct, Llama-3-70B-Instruct, Llama-3.1-8B-Instruct, and Llama-3.1-70B-Instruct. For these nine pairs we re-ran both the base and adapted checkpoints on all benchmarks at the same 2,048-token budget, greedy decoding, and seed as the primary arm, changing only the prompt wrapping: the identical question, choice block, and answer instruction were rendered through each checkpoint's native chat template as a single user turn. Each adapted checkpoint shares its base's chat template exactly, confirming that the contrast is not confounded by template differences within a pair.

\paragraph{Result.} Under native chat templates, \MAAP appropriate abstention decreased in seven of nine pairs with a median change of $-2.88$ pp (base-clustered 95\% CI: $-10.43$ to $+0.10$ pp; Wilcoxon $p=0.074$; base-clustered percentile-sign $p=0.109$). The same nine pairs under the primary completion protocol gave a median change of $-1.89$ pp with eight of nine decreasing (95\% CI: $-8.44$ to $-0.70$ pp; $p=0.012$; $p_{\mathrm{cluster}}=0.011$). The direction of the effect is therefore reproduced under native prompting, with a larger median magnitude.

\paragraph{Within-pair format differences.} We also wish to test directly whether prompt format changes the measured effect for the relevant pairs. The within-pair difference between the native-template and completion effects has a median of $+0.79$ pp (95\% CI: $-3.67$ to $+5.06$ pp; Wilcoxon $p=0.426$; $p_{\mathrm{cluster}}=0.461$), which is small relative to the effect itself and statistically indistinguishable from zero. %
In fact, no outcome we examined showed significant format moderation; the largest was \CDC accuracy at $-6.25$ pp ($p=0.113$, $p_{\mathrm{cluster}}=0.112$).

\paragraph{Discussion.} While this sensitivity analysis confirms that the choice of completion template did not instrument the effects reported in the paper, there are two limitations of note. First, the native-template arm does not reach significance on its own-- its clustered interval includes zero at the margin. With nine pairs over four base clusters, this reflects the power available in the small subset rather any substantive evidence against the main reported effect of the work. Second, the nine eligible pairs are not a random sample of the cohort. They are restricted to the Llama-3 and Llama-3.1 Instruct lineages, and their completion-arm median change of $-1.89$ pp is smaller than the $-3.48$ pp observed across all 25 pairs. Hence, the subset is not representative of the full-cohort effect size. Taken together, the evidence supports the conclusion that the primary result does not stem from the common completion protocol, though the effect of the native-template itself would require more model pairs to resolve to a satisfactory level of significance.

This analysis varies prompt \emph{format} while holding prompt \emph{content} fixed, ruling out a major potential source of the effects we report in this work. It does not, however, evaluate publisher-recommended system prompts. Those differ across releases in content that bears directly on the measured outcome; for instance, one adapted checkpoint's documented system prompt instructs the model not to share information it is uncertain of, while another's mandates a specific answer format. On the other hand, none of the four base checkpoints documents a substantive system prompt. Pairing each release with its own system prompt would therefore confound the adaptation contrast with each publisher's prompt, changing the evaluation from one measuring the effect of adaptation under a fixed protocol to one measuring the capability, caution, and commitment as induced by adaptation and a system prompt. We leave that comparison to future work.

\begin{table*}[htbp]
\centering
\caption{Per-pair change in \MAAP \ appropriate abstention (percentage points) under the primary completion protocol and under each checkpoint's native chat template, for the nine pairs with a chat-capable base checkpoint. Negative values indicate reduced appropriate abstention after adaptation.}
\label{tab:native-template-pairs}
\small
\begin{tabular}{llrr}
\toprule
Adapted checkpoint & Base checkpoint & Completion & Native chat \\
\midrule
Aloe-Beta-70B        & Llama-3.1-70B-Instruct & $-0.70$  & $+0.10$  \\
Aloe-Beta-8B         & Llama-3.1-8B-Instruct  & $-8.44$  & $-3.38$  \\
HuatuoGPT-o1-70B     & Llama-3.1-70B-Instruct & $-0.89$  & $+3.28$  \\
HuatuoGPT-o1-8B      & Llama-3.1-8B-Instruct  & $-1.89$  & $-0.60$  \\
Med42-v2-70B         & Llama-3-70B-Instruct   & $-3.48$  & $-3.18$  \\
Med42-v2-8B          & Llama-3-8B-Instruct    & $-21.25$ & $-10.43$ \\
Meerkat-8B           & Llama-3-8B-Instruct    & $-5.66$  & $-13.01$ \\
OpenBioLLM-70B       & Llama-3-70B-Instruct   & $-1.49$  & $-2.18$  \\
Palmyra-Med-70B-32K  & Llama-3-70B-Instruct   & $+0.79$  & $-2.88$  \\
\midrule
Median               &                        & $-1.89$  & $-2.88$  \\
\bottomrule
\end{tabular}
\end{table*}

\subsection{Probability Capture and Calibration-Metric Sensitivity} \label{app:probability-capture}
Probability metrics are computed on the jointly captured population, meaning the items for which both roles yield a usable choice distribution at the answer position. The validity of this choice depends on how often a choice falls outside the saved top-20 tokens and how far the reported calibration change depends on the binning of the expected calibration error.

\paragraph{Capture.} On \MA, the median pairwise joint capture rate is 100.0\% (96.43\% of aligned items pooled across pairs); on \CD, the median pairwise rate is 93.62\%. Omission of a declared choice from the top-20 list is rare for base checkpoints (median share of captured items with at least one option missing is 0.12\% on \MA\ and 0.00\% on \CD) and remains uncommon, though more frequent, for adapted checkpoints (4.27\% and 1.08\%). Candidate omission is therefore not a material source of the reported effects. The mass lying outside the declared choice set before renormalization is substantial, however, with a median of 41.85\% for base and 60.80\% for adapted checkpoints on \MA. Appendix~\ref{app:full-vocab-rescoring} addresses that second concern directly by rescoring over the full vocabulary.

\paragraph{Binning.} The \MANA\ calibration increase does not depend on the ten-bin choice. Median pair-level ECE change is $+0.0399$ at ten bins (Wilcoxon $p=0.0042$; $p_{\mathrm{cluster}}=0.060$), $+0.0378$ at fifteen bins ($p=0.0051$; $p_{\mathrm{cluster}}=0.014$), and $+0.0345$ at twenty bins ($p=0.0061$; $p_{\mathrm{cluster}}=0.033$). The proper scoring rules are less decisive-- the Brier change has a median of $+0.0331$ ($p=0.325$) and the negative log-likelihood change a median of $+0.2009$ ($p=0.096$), neither of which meets our significance criterion. We therefore report the ECE result as robust to binning and treat the Brier and log-likelihood changes as inconclusive rather than as corroborating evidence.

\subsection{Full-Vocabulary Rescoring of Answer-Token Confidence} \label{app:full-vocab-rescoring}
The answer-token quantities in Section~\ref{sec:commitment} are reconstructed from the top-20 token logprobs saved at each generated position, then renormalized over the declared choice set.
Two questions follow:
\begin{enumerate}
    \item Whether choice surfaces fall outside the saved top 20, and
    \item Whether re-normalizing away the remaining mass changes the reported effect.
\end{enumerate}
The capture diagnostics in Appendix~\ref{app:probability-capture} addressed the first.
Here we address the second by recomputing the distribution exactly over the entire vocabulary, concluding that \textbf{exact scoring does not change the direction or the ordering of the effect.}

Across ten pairs spanning fifteen checkpoints of 4 to 8 billion parameters, replacing the choice-normalized top-20 quantity with the exact, non-renormalized full-vocabulary probability leaves the sign of the shared-error commitment change unaltered in every pair, and preserves their rank ordering almost exactly (Spearman $\rho=0.99$, Pearson $r=0.99$).
The per-pair pattern also tracks the values obtained from the original runs ($\rho=0.99$); the only two sign differences arise in pairs whose effect lies within 0.05 of zero, where the smaller item sample used here cannot resolve a direction.
The conclusions of Section~\ref{sec:commitment} are therefore unchanged under exact scoring.

The one observed change is a modest one in magnitude. Renormalization inflates the measured commitment change in all ten pairs, by a median of 0.7 points and at most 8.4 points (Wilcoxon $p=0.002$).
Against the cohort effect of 11.2 points this is a median bias of 6.6\%, reaching 75\% in the most affected pair.
In interpreting this change, we observe the probability mass lying outside the choice set does not shift systematically with adaptation (median $+2.1$ points, $p=0.432$) and increases in eight of ten pairs, so the results do not support the hypothesis that increased commitment reflects probability migrating onto answer-letter tokens from outside the top-20 token set.

We performed this additional experiment by teacher-forcing the saved generated prefix for each model response and reading the full-vocabulary distribution at the position that predicts the saved answer token, making the comparison item-for-item by construction.
The residual disagreements between the original run and this experiment are small-- where the two implementations select different tokens, the median gap between the top two choices is 0.03, against 0.60 where they agree.
The exact renormalized distribution matches the archived object to within 0.007 mean absolute difference for every checkpoint.

We perform this follow-on experiment on the 4-8B model subset to validate the measurement; it should not be considered an independent test of the effect.
The commitment change is positive in seven of ten pairs with a median of $+7.8$ points, consistent with the full cohort (median $+11.2$ points, positive in 22 of 25), but it is not individually significant here ($p=0.193$), due to the lower power. Expected calibration error is likewise uninformative at 150 \texttt{NA} items per checkpoint.

\subsection{Additional Results}
\label{app:add-figures}

\begin{figure*}[htbp]
    \centering
    \includegraphics[width=\textwidth]{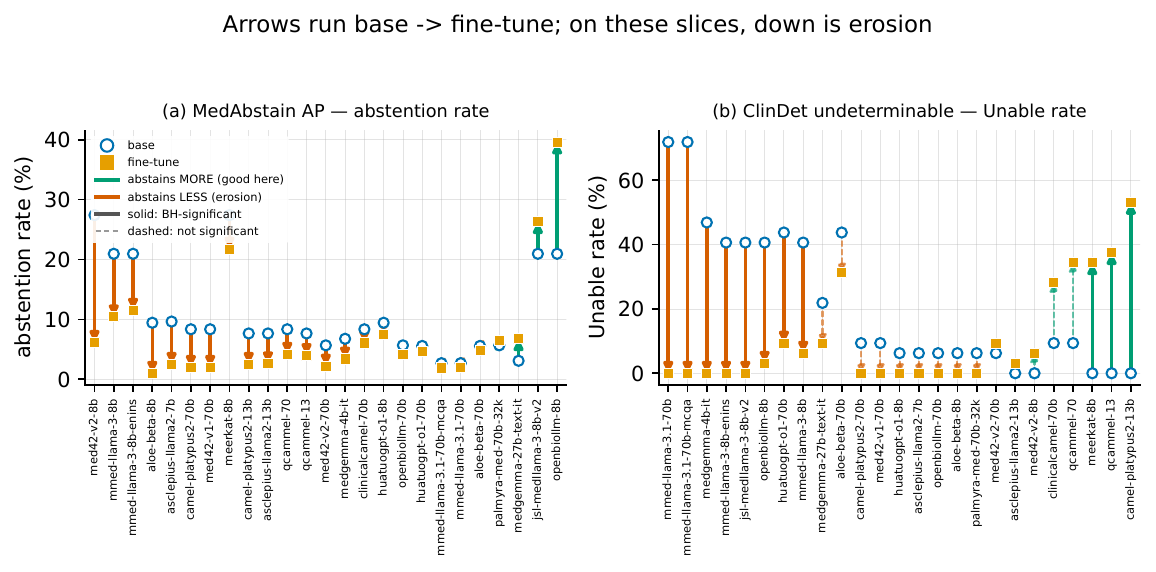}
    \caption{{Abstention (``Unable'') rate by model role for \MAAP\ (left) and \CDIU\ (right). Solid lines connect base--fine-tune pairs with a BH-significant change; dashed lines denote non-significant pairs. A fine-tuned rate below its paired base rate indicates erosion of appropriate caution.}}
    \label{fig:abstention_levels}
\end{figure*}

The paired model-level abstention rates underlying the main-text cohort result are shown in \cref{fig:abstention_levels}.

Since the public release of MedAbstain does not expose the item-level human-validation labels, we cannot judge model behavior on only the questions deemed unanswerable by a clinician. In aggregate, clinicians judged 22.45\% of labeled perturbed cases answerable, whereas the median adapted checkpoint answers 95.83\% of AP items. Therefore, reduced abstention on the answerable share cannot fully explain the loss in caution observed, but a more careful analysis at the individual question level is not possible at this time.

\begin{figure*}[htbp]
    \centering
    \includegraphics[width=0.75\textwidth]{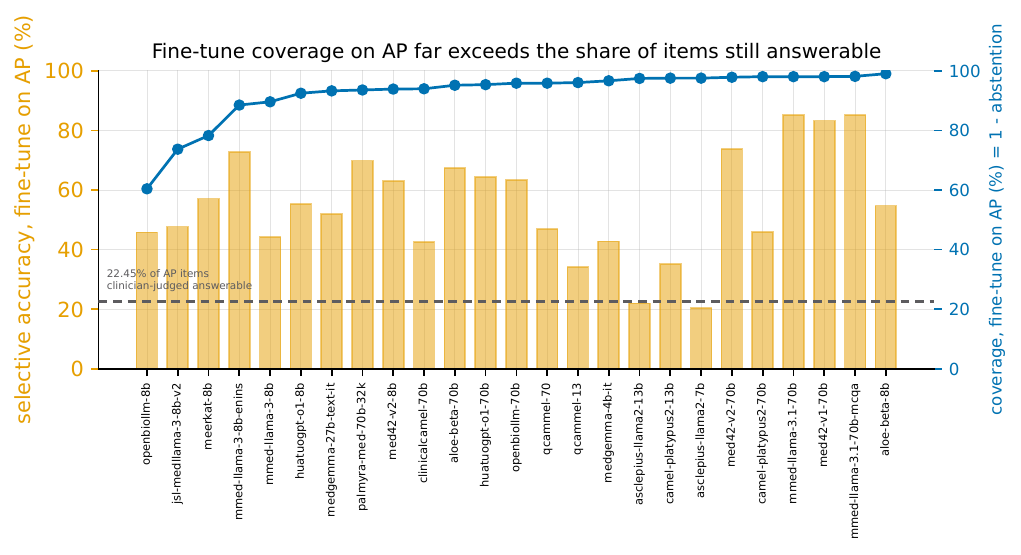}
    \caption{{Selective accuracy versus coverage for fine-tuned models on \MAAP. The reference line marks the benchmark annotation's estimated answerable share; every fine-tuned model's coverage exceeds this share. This aggregate comparison does not identify which items drive pair-level abstention changes.}}
    \label{fig:ap_answerable_rebuttal}
\end{figure*}

\begin{figure*}[htbp]
    \centering
    \includegraphics[width=\textwidth]{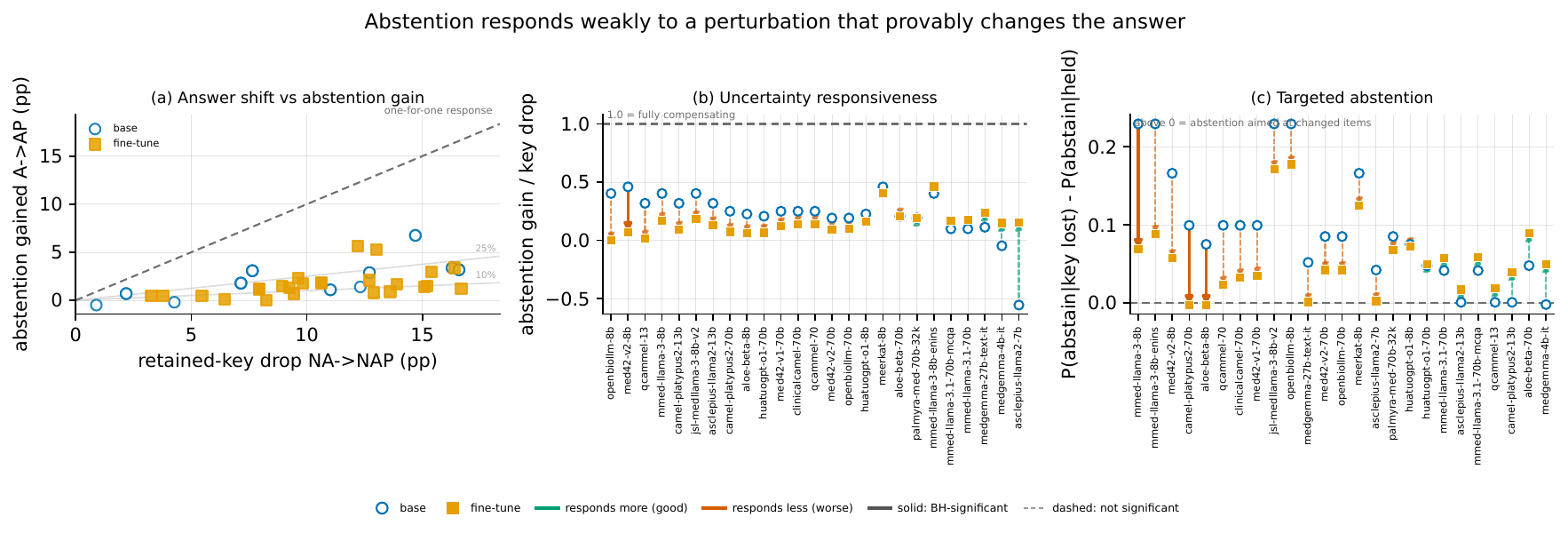}
    \caption{{Responsiveness of MedAbstain abstention to evidence removal. Left: retained-key drop from \MANA\ to \MANAP\ versus abstention gain from \MAA\ to \MAAP, by role. Center: responsiveness ratio $\mathrm{RR}_r$ per model; a value of one indicates fully compensating abstention. Right: targeting statistic $T_r$, the excess \MAAP\ abstention on items whose retained answer changed under perturbation relative to items whose answer did not change.}}
    \label{fig:responsiveness_targeting}
\end{figure*}

The role-level diagnostics in \cref{fig:responsiveness_targeting} refine the informedness result. Evidence removal reduced retained-key agreement by a median of 7.65 pp for base roles and 10.63 pp for fine-tuned roles, whereas abstention increased by only 2.88 pp and 1.39 pp, respectively. Median within-role responsiveness ratios were 0.250 and 0.140, and none of the 50 estimable roles reached $\mathrm{RR}=1$; thus, neither role fully converted lost answer support into abstention. Fine-tuning also reduced item-level targeting in the unclustered analysis, but this effect did not survive base-clustered inference.

\begin{figure*}[htbp]
    \centering
    \includegraphics[width=\textwidth]{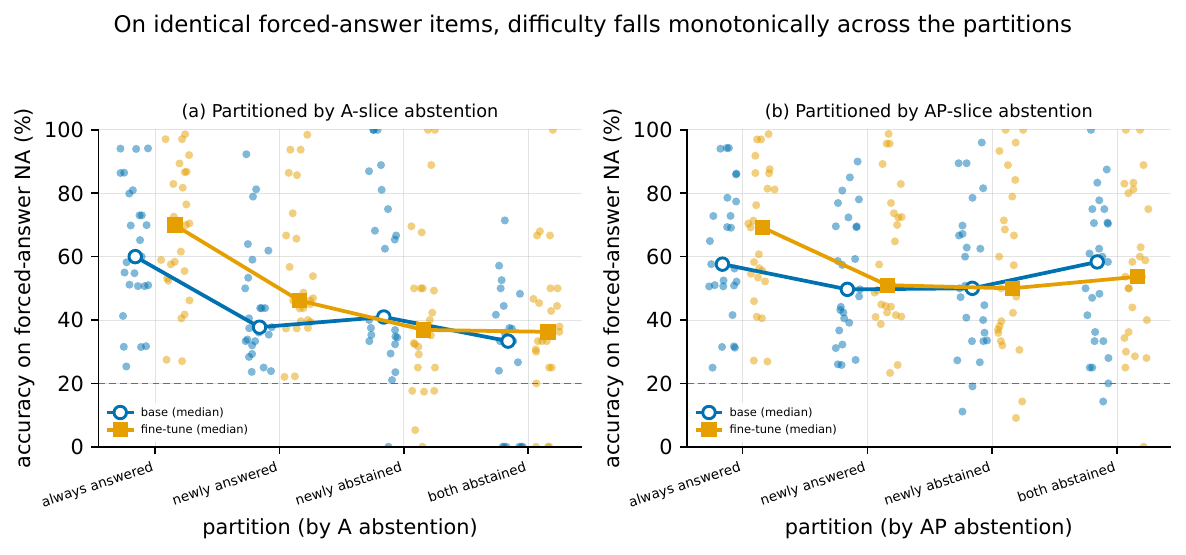}
    \caption{{Median accuracy on no-abstention \MANA\ questions, computed within the coverage-transition partitions defined on \MAA\ (left) and \MAAP\ (right): always answered, newly answered, newly abstained, and both abstained.}}
    \label{fig:na_partition_accuracy}
\end{figure*}

When the same partitions were transferred to \MANA, where neither model could abstain, no within-partition fine-tuned-versus-base accuracy change survived multiplicity correction (\cref{fig:na_partition_accuracy}). The robust result is therefore that coverage changes concentrate on harder questions, not that fine-tuning uniformly reduces knowledge within a particular partition.

\begin{figure*}[htbp]
    \centering
    \includegraphics[width=\textwidth]{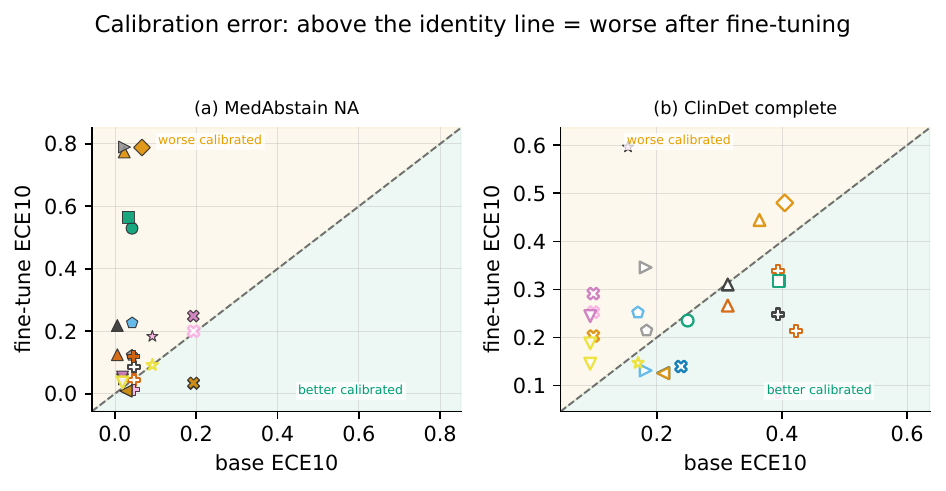}
    \caption{{Fine-tune versus base ECE10 for each model pair on \MANA\ and \CDC. Points above the identity line indicate worse calibration after fine-tuning.}}
    \label{fig:ece_identity}
\end{figure*}

\MANA\ ECE10 increased by 0.040, or 99.8\% relative to the base value (\cref{fig:ece_identity}). The increase survived correction in both the ten-test commitment-and-calibration (\textbf{H4}) family ($q_{\mathrm{BH}}=0.010$) and the broader 15-claim hypothesis set ($q_{\mathrm{BH}}=0.009$), as well as base-clustered inference ($p_{\mathrm{cluster}}=0.040$). Confidence AUROC changed by only $-0.015$ ($q_{\mathrm{BH}}=0.508$, $p_{\mathrm{cluster}}=0.873$), indicating no robust improvement in the ability of confidence to distinguish correct from incorrect answers. Corresponding ECE10 comparisons for \MAA, \MANAP, and \MAAP\ are shown in Appendix Figure~\ref{fig:ece_identity_appendix}.

On \CDC, ECE10 did not change reliably. Entropy and shared-wrong confidence survived the unclustered commitment and calibration (\textbf{H4}) correction but attenuated under base clustering, and AUROC no longer survives correction; only the confidence increase also holds under base clustering. The probability-distribution changes are therefore most robust on \MANA.

\begin{figure*}[htbp]
    \centering
    \includegraphics[width=\textwidth]{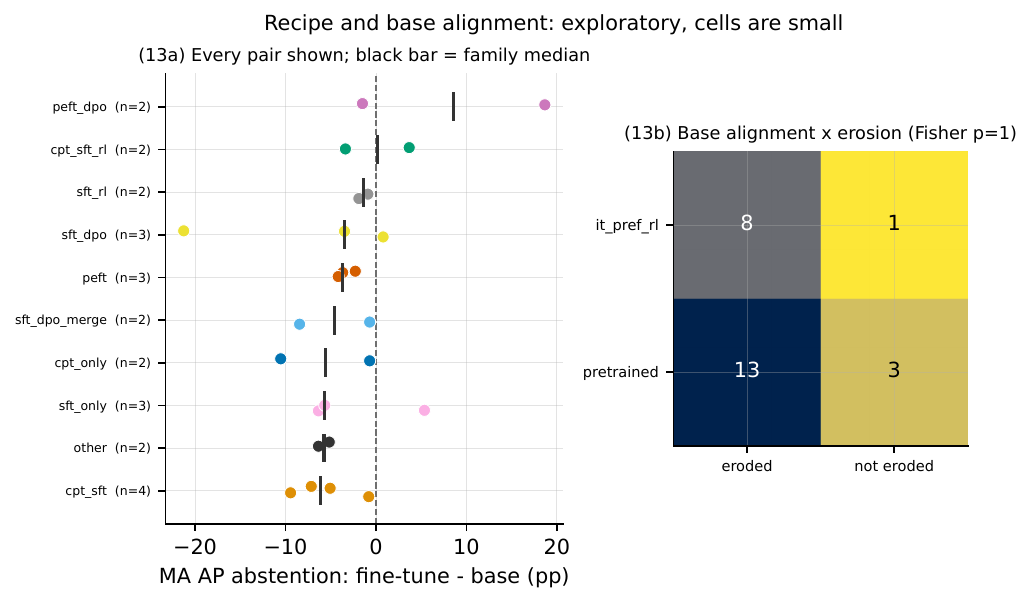}
    \caption{{Left: pair-level change in \MAAP\ abstention (pp), grouped by broad fine-tuning recipe family; the black bar marks the family median. Right: base-model alignment against the erosion pattern. No recipe-family omnibus test survived multiplicity correction, and the alignment contrast is not significant.}}
    \label{fig:moderators_recipe}
\end{figure*}

\begin{figure*}[htbp]
    \centering
    \includegraphics[width=0.6\textwidth]{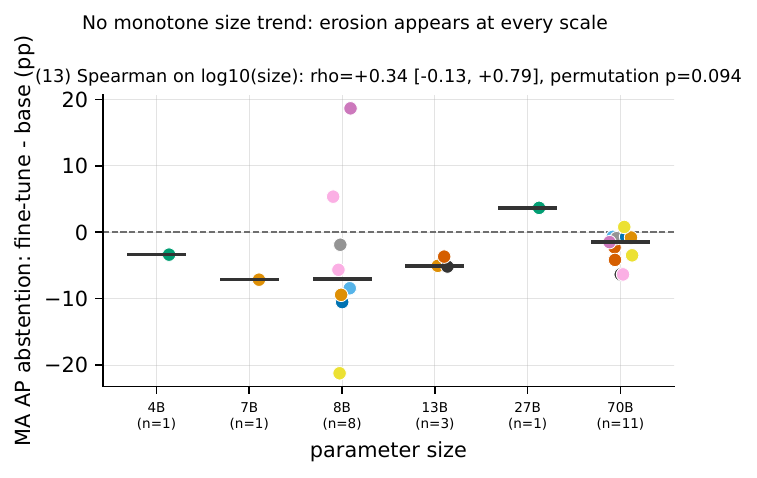}
    \caption{{Pair-level change in \MAAP\ abstention (pp) against base-model parameter count. The Spearman correlation on $\log_{10}$(size) is positive but not statistically significant ($\rho=0.34$, permutation $p=0.094$).}}
    \label{fig:size_scaling}
\end{figure*}

We explored whether heterogeneity was associated with model size, base-model alignment, fine-tuning strength, or broad recipe family. None provided a stable explanation after multiplicity correction. Recipe-family omnibus tests were non-significant for \MAAP\ abstention ($p=0.529$; \cref{fig:moderators_recipe}), \CDIU\ caution ($p=0.104$), and the newly answered accuracy gap ($p=0.056$). Neither base-model alignment nor fine-tuning strength was significantly associated with the primary outcomes after correction, and guardrail erosion showed no statistically significant monotonic relationship with parameter count (\cref{fig:size_scaling}).

\begin{figure*}[htbp]
    \centering
    \includegraphics[width=0.7\textwidth]{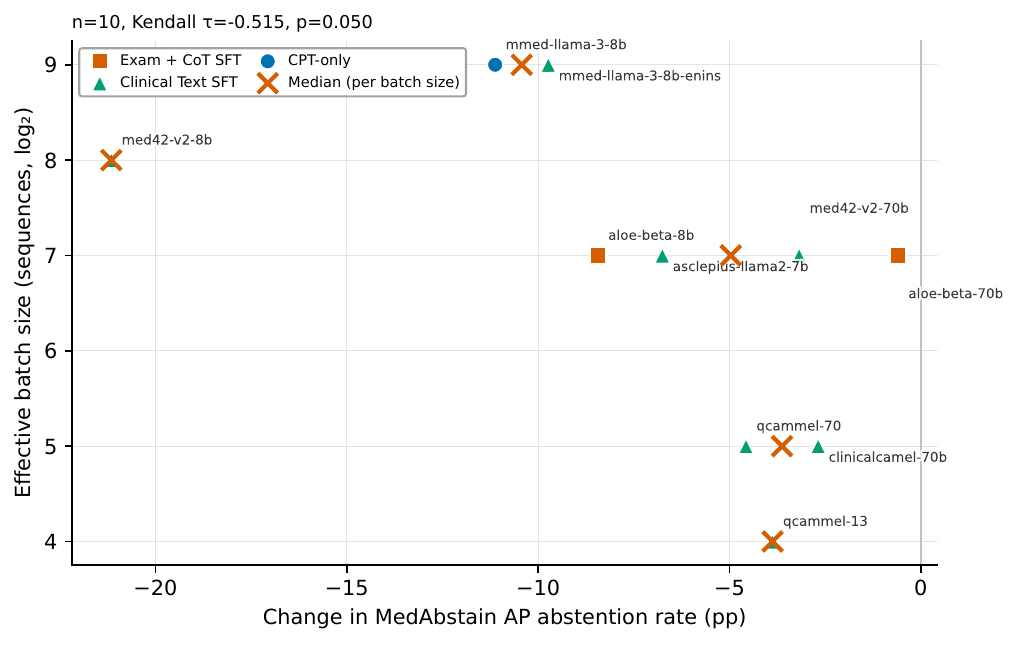}
    \caption{{Change in \MAAP\ abstention (pp) against effective training batch size for the subset with disclosed batch size ($n=10$). The result is exploratory.}}
    \label{fig:recipe_batch_kendall}
\end{figure*}

\begin{figure*}[htbp]
    \centering
    \includegraphics[width=\textwidth]{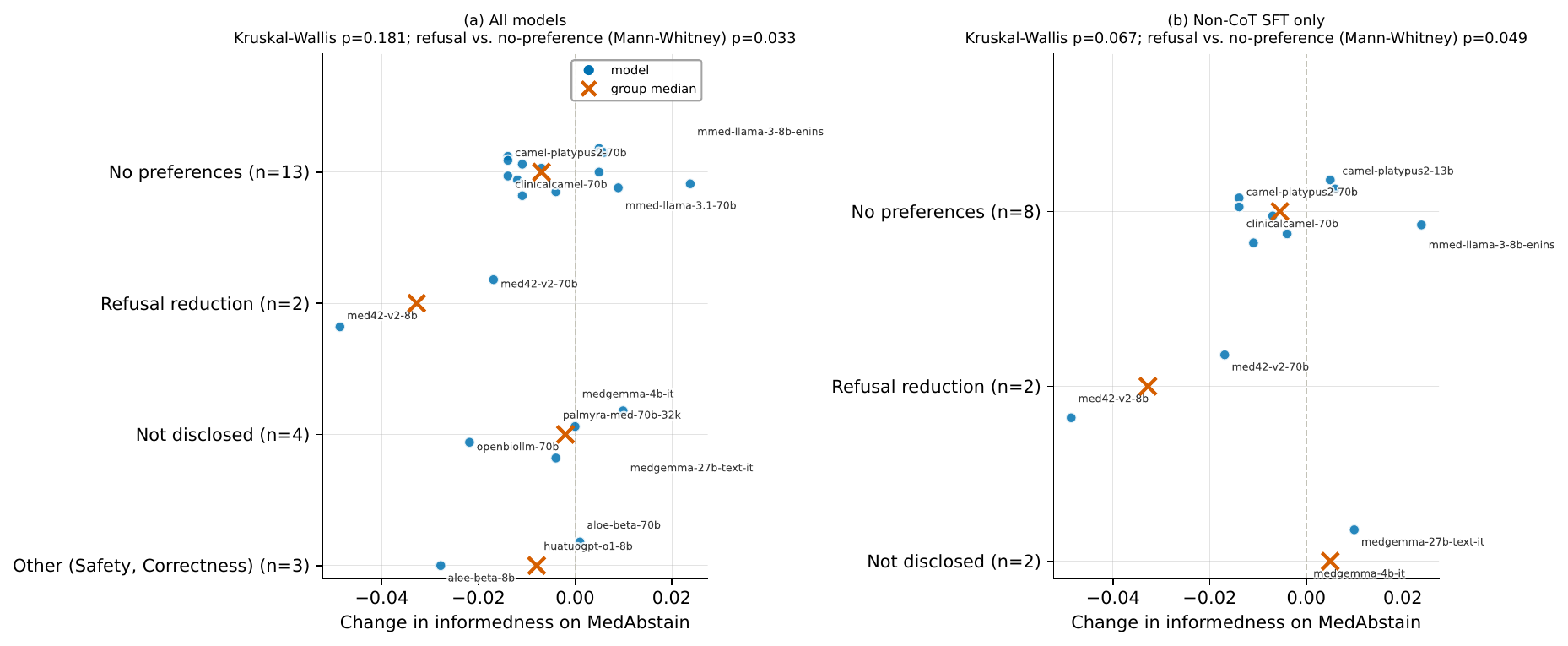}
    \caption{{Change in \MA\ informedness grouped by disclosed preference-optimization objective. The two refusal-reduction pairs come from one release family, and omnibus tests are not significant.}}
    \label{fig:recipe_preference_informedness}
\end{figure*}

\begin{figure*}[htbp]
    \centering
    \includegraphics[width=\textwidth]{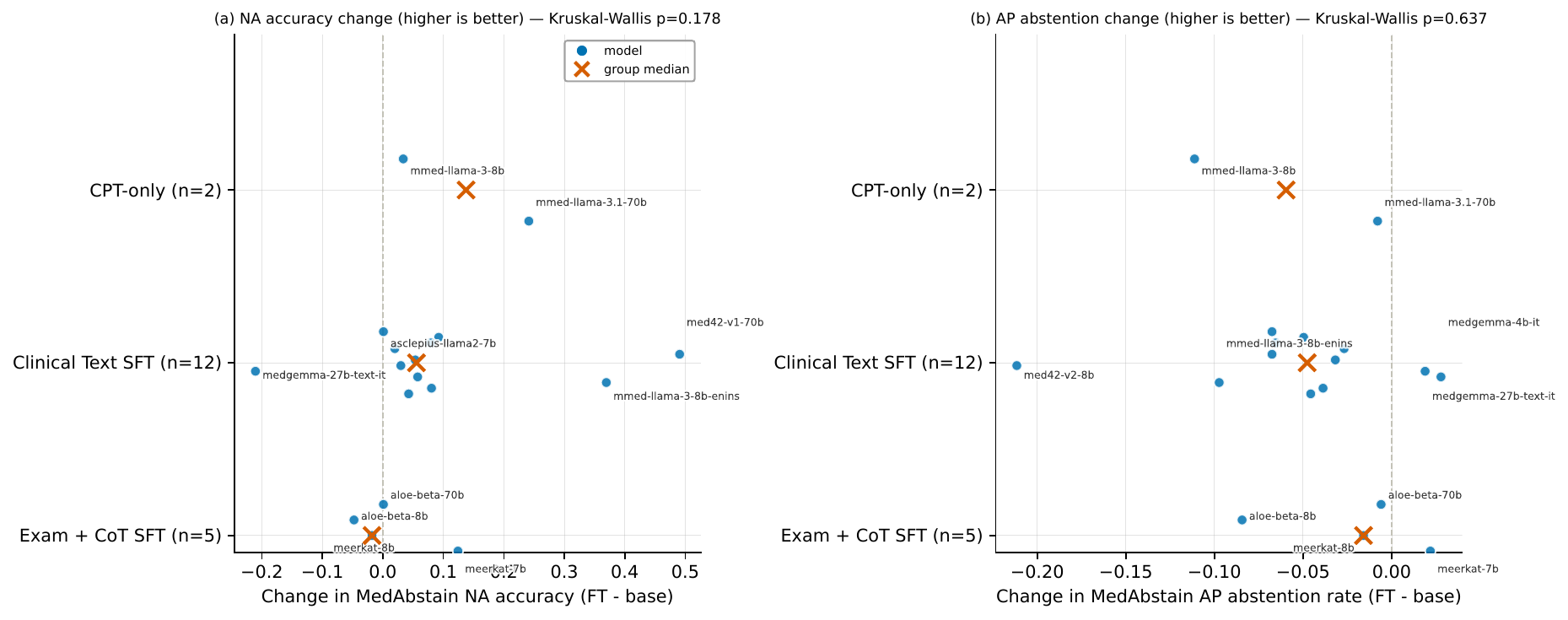}
    \caption{{Change in \MANA\ accuracy and \MAAP\ abstention grouped by broad fine-tuning data content. Neither omnibus test survives correction.}}
    \label{fig:recipe_accuracy_caution}
\end{figure*}

Several small recipe groups showed large directional changes, including models using continued pre-training, preference optimization, refusal-reduction objectives (\cref{fig:recipe_preference_informedness}), or larger effective batch sizes (\cref{fig:recipe_batch_kendall}). The two continued-pretraining-only models showed the largest median capability gain among the broad data-content groups, but neither the capability nor caution omnibus test survived correction (\cref{fig:recipe_accuracy_caution}). However, recipe cells generally contained only two to four pairs, and these characteristics were confounded with model family, parameter count, data composition, and training configuration. We therefore treat these patterns as hypotheses for a controlled follow-up study rather than evidence that any recipe component caused guardrail erosion.

\clearpage
\subsection{Benchmark Landscape}
\label{app:bench-all}
We considered benchmarks for the audit along two dimensions-- the \emph{guardrail} they probe (knowledge-centric epistemic behavior versus value-centric ethical refusal) and their \emph{domain} (clinical versus general). Knowledge-centric benchmarks are further grouped by the concept tested: capability (correctness), hallucination (faithfulness), and determinacy/abstention (recognizing when evidence is insufficient). Table \ref{tab:benchmarks} contains the list of benchmarks we considered.
\subsubsection{Benchmarks Used in the Study}
\label{app:bench-used}
\paragraph{MedAbstain (\MA).} MedAbstain, introduced by \citet{machchaetal2026knowinga}, provides an effective way to measure models' ability to abstain when presented with idiosyncratic uncertainty (namely, insufficient patient information for making a diagnosis). The benchmark takes a multiple choice medical knowledge question, e.g., from MedQA (\cite{jinetal2021what}), and uses it to create four variants of questions using two axes of variation-- whether key evidence is ablated (perturbed versus original) and whether an abstention choice is offered.
The four variants are hence No-Abstention (\MANA), Abstention (\MAA), No-Abstention-Perturbed (\MANAP), and Abstention-Perturbed (\MAAP).

These four variants aim to triangulate a model's behavior when presented with incomplete information, and further isolate appropriate abstention (in the face of uncertainty).
In particular, we are most interested in the model's abstention rate on the \MAAP\ questions, which warrant abstention in the vast majority of cases (\citet{machchaetal2026knowinga} report that human annotators judged abstention was appropriate in 38 of 49 \MAAP\ questions, leaving 22\% still answerable after perturbation).
For the purposes of our evaluations, we focus on the publicly accessible MedQA-USMLE-derived half of \MA, consisting of 1,007 items per variant. Each question is accompanied by 5 multiple-choice options, plus the additional abstain option for the \MAA\ and \MAAP\ variants.

\paragraph{ClinDet-Bench (\CD).}
ClinDet-Bench \citep{watanabeetal2026clindetbencha} tests a concept related to but distinct from \MA\ -- whether a model is able to judge if missing information is material to the task at hand. The protocol uses score-based diagnosis frameworks (e.g., CHADS2 \cite{gageetal2001validation}), which assigns point scores based on the presence of risk factors (e.g., age 75 years or older, history of stroke). \CDI\ instances have some risk factors masked, compared to \CDC\ instances which present all risk factors. A \CDI\ instance is \textbf{Determinable} if every possible value of the missing risk factors results in the same diagnosis; it is \textbf{Undeterminable} otherwise (note that by construction, all \CDC\ instances are Determinable).
\begin{table*}[htbp]
\centering
\caption{Benchmarks considered for the audit. The two
benchmarks used for our primary results are marked with a \ourbench. Our study targets the
clinical, knowledge-centric quadrant, which leaves us with the two benchmarks considered
in our study.}
\label{tab:benchmarks}
\resizebox{\textwidth}{!}{%
\begin{tabular}{@{}L{3.4cm} L{2.2cm} L{3.9cm} L{4.6cm} c@{}}
\toprule
\textbf{Benchmark} & \textbf{Domain} & \textbf{Concept} & \textbf{Description} & \textbf{Ref.} \\
\midrule
\multicolumn{5}{@{}l}{\textit{Knowledge-centric guardrails}} \\
\midrule
MedQA (USMLE) & Clinical & Capability & 4/5-option board-exam MCQA& \cite{jinetal2021what} \\
MedMCQA & Clinical & Capability & Large multi-subject Indian medical entrance MCQA & \cite{paletal2022medmcqa} \\
PubMedQA & Clinical & Capability & Yes/no/maybe QA on PubMed abstracts & \cite{jinetal2019pubmedqa} \\
EHRNoteQA & Clinical & Capability (notes) & QA requiring synthesis across MIMIC-IV discharge notes & \cite{kweonetal2024ehrnoteqaa} \\
Med-HALT & Clinical & Hallucination / reasoning & Reasoning + memory hallucination tests & \cite{paletal2023medhalta} \\
MedHallu & Clinical & Calibration / hallucination & Detecting fabricated answers derived from PubMedQA & \cite{panditetal2025medhallu} \\
MedAbstain \ourbench & Clinical & Abstention / calibration & Selective-prediction MCQA with context ablation & \cite{machchaetal2026knowinga} \\
ClinDet-Bench \ourbench & Clinical & Determinacy & Determinability on diagnostic scoring frameworks & \cite{watanabeetal2026clindetbencha} \\
MediQ & Clinical & Information-seeking & Interactive diagnosis capability; asking for missing information & \cite{lietal2024mediq} \\
\addlinespace
MMLU & General & Capability & 57-subject multitask knowledge MCQA & \cite{hendrycksetal2021measuring} \\
TruthfulQA & General & Calibration / truthfulness & Resistance to imitating common human falsehoods & \cite{linetal2022truthfulqa} \\
SQuAD 2.0 & General & Abstention & Extractive QA with unanswerable questions to reject & \cite{rajpurkaretal2018know} \\
\midrule
\multicolumn{5}{@{}l}{\textit{Value-centric (ethics) guardrails}} \\
\midrule
MedSafetyBench & Clinical & Refusal & Harmful medical requests mapped to AMA ethical principles & \cite{hanetal2024medsafetybencha} \\
SOSBench (med.) & Clinical & Refusal & Regulation-grounded harmful requests containing medical questions & \cite{jiangetal2025sosbench} \\
SciSafeEval (med.) & Clinical & Refusal & Refusal vs.\ helpfulness on malicious scientific instructions & \cite{lietal2024scisafeeval} \\
\addlinespace
AdvBench & General & Refusal & Harmful behaviors targeted by transferable adversarial prompts & \cite{zouetal2023universal} \\
HarmBench & General & Refusal & Standardized automated red-teaming and robust refusal & \cite{mazeikaetal2024harmbench} \\
StrongREJECT & General & Refusal & Jailbreak effectiveness with a calibrated harmfulness scorer & \cite{soulyetal2024strongreject} \\
\bottomrule
\end{tabular}%
}
\end{table*}

\subsection{Models Evaluated}

\Cref{tab:models} provides details on the 11 base foundation models considered in our study, while \cref{tab:finetuned-models} describes the 25 fine-tuned models adapted for the biomedical and clinical domains.
\begin{table*}[htbp]
\centering
\caption{Base checkpoints for the 25 domain-adapted models considered in our study. ``Type'' refers to pretrained-only
(PT) checkpoints versus instruction/preference-tuned (IT) checkpoints.
}
\label{tab:models}
\resizebox{\textwidth}{!}{%
\begin{tabular}{@{}L{3.4cm} L{1.7cm} c c L{7.6cm} c@{}}
\toprule
\textbf{Base checkpoint} & \textbf{Family} & \textbf{Params} & \textbf{Type} & \textbf{HF repository} & \textbf{Ref.} \\
\midrule
Llama-2-7B          & Llama-2   & 7B  & PT & \texttt{meta-llama/Llama-2-7b-hf}            & \cite{touvronetal2023llamaa} \\
Llama-2-13B         & Llama-2   & 13B & PT & \texttt{meta-llama/Llama-2-13b-hf}           & \cite{touvronetal2023llamaa} \\
Llama-2-70B         & Llama-2   & 70B & PT & \texttt{meta-llama/Llama-2-70b-hf}           & \cite{touvronetal2023llamaa} \\
Llama-3-8B          & Llama-3   & 8B  & PT & \texttt{meta-llama/Meta-Llama-3-8B}          & \cite{grattafiorietal2024llama} \\
Llama-3-8B-Instruct & Llama-3   & 8B  & IT & \texttt{meta-llama/Meta-Llama-3-8B-Instruct} & \cite{grattafiorietal2024llama} \\
Llama-3-70B-Instruct& Llama-3   & 70B & IT & \texttt{meta-llama/Meta-Llama-3-70B-Instruct}& \cite{grattafiorietal2024llama} \\
Llama-3.1-8B-Instruct  & Llama-3.1 & 8B  & IT & \texttt{meta-llama/Llama-3.1-8B-Instruct}  & \cite{grattafiorietal2024llama} \\
Llama-3.1-70B          & Llama-3.1 & 70B & PT & \texttt{meta-llama/Llama-3.1-70B}          & \cite{grattafiorietal2024llama} \\
Llama-3.1-70B-Instruct & Llama-3.1 & 70B & IT & \texttt{meta-llama/Llama-3.1-70B-Instruct} & \cite{grattafiorietal2024llama} \\
Gemma-3-4B-PT       & Gemma-3   & 4B  & PT & \texttt{google/gemma-3-4b-pt}                & \cite{gemmateam2025gemma3} \\
Gemma-3-27B-PT      & Gemma-3   & 27B & PT & \texttt{google/gemma-3-27b-pt}               & \cite{gemmateam2025gemma3} \\
\bottomrule
\end{tabular}%
}
\end{table*}

\begin{table*}[htbp]
\centering
\caption{The 25 biomedical domain-adapted models in the study, grouped by base family. \emph{Recipe} lists the disclosed post-training stages in order (CPT: continued pre-training; SFT: supervised
fine-tuning; DPO: direct preference optimization; RL: reinforcement learning;
QLoRA: quantized low-rank adaptation; MERGE: weight merging).
\emph{\MA\ cont.}\ is the MedAbstain contamination flag (\yes\ =
confirmed reuse of MedQA/USMLE-derived items; \plaus\ = plausible but unconfirmed;
no = clean). Style records the model release's documented chat or completion format; all benchmark evaluations used the common completion protocol described in the main text. The Llama-3 lineage shown for Palmyra-Med is inferred from the architecture of its disclosed Palmyra-x base rather than explicitly stated by its model card.}
\label{tab:finetuned-models}

\resizebox{\textwidth}{!}{%
\begin{tabular}{@{}L{4.5cm} c L{2.7cm} L{3.4cm} c c c c@{}}
\toprule
\textbf{Domain-adapted model} & \textbf{Params} & \textbf{Base checkpoint} & \textbf{Recipe} & \textbf{Stg.} & \textbf{Style} & \textbf{\MA\ cont.} & \textbf{Ref.} \\
\midrule
\multicolumn{8}{@{}l}{\textit{Llama-2 family}} \\
\midrule
Asclepius-Llama2-7B  & 7B  & Llama-2-7B  & CPT + SFT   & 2 & compl. & no     & \cite{kweonetal2024asclepius} \\
Asclepius-Llama2-13B & 13B & Llama-2-13B & CPT + SFT   & 2 & compl. & no     & \cite{kweonetal2024asclepius} \\
Camel-Platypus2-13B  & 13B & Llama-2-13B & MERGE       & 1 & compl. & \plaus & \cite{leeetal2023platypus} \\
qCammel-13           & 13B & Llama-2-13B & QLoRA       & 1 & compl. & \plaus & \cite{tomaetal2023clinical} \\
Camel-Platypus2-70B  & 70B & Llama-2-70B & MERGE       & 1 & compl. & \plaus & \cite{leeetal2023platypus} \\
ClinicalCamel-70B    & 70B & Llama-2-70B & QLoRA       & 1 & compl. & \yes   & \cite{tomaetal2023clinical} \\
Med42-v1-70B         & 70B & Llama-2-70B & SFT         & 1 & compl. & \plaus & \cite{christopheetal2024med42} \\
qCammel-70           & 70B & Llama-2-70B & QLoRA       & 1 & compl. & \plaus & \cite{tomaetal2023clinical} \\
\midrule
\multicolumn{8}{@{}l}{\textit{Llama-3 family}} \\
\midrule
JSL-MedLlama-3-8B-v2 & 8B  & Llama-3-8B          & SFT         & 1 & compl. & \plaus & \cite{johnsnowlabs2024jslmedllama} \\
MMed-Llama-3-8B      & 8B  & Llama-3-8B          & CPT         & 1 & compl. & \plaus & \cite{qiuetal2024building} \\
MMed-Llama-3-8B-EnIns& 8B  & Llama-3-8B          & CPT + SFT   & 2 & chat   & \yes   & \cite{qiuetal2024building} \\
OpenBioLLM-8B        & 8B  & Llama-3-8B          & QLoRA + DPO & 2 & chat   & \plaus & \cite{paletal2024openbiollm} \\
Med42-v2-8B          & 8B  & Llama-3-8B-Instruct & SFT + DPO   & 2 & chat   & \plaus & \cite{christopheetal2024med42v2} \\
Meerkat-8B           & 8B  & Llama-3-8B-Instruct & SFT         & 1 & chat   & \plaus & \cite{kimetal2025small} \\
OpenBioLLM-70B       & 70B & Llama-3-70B-Instruct& QLoRA + DPO & 2 & chat   & \plaus & \cite{paletal2024openbiollm} \\
Med42-v2-70B         & 70B & Llama-3-70B-Instruct& SFT + DPO   & 2 & chat   & \plaus & \cite{christopheetal2024med42v2} \\
Palmyra-Med-70B-32K  & 70B & Llama-3-70B-Instruct (inferred)& SFT + DPO   & 2 & chat   & \plaus & \cite{writer2024palmyramed} \\
\midrule
\multicolumn{8}{@{}l}{\textit{Llama-3.1 family}} \\
\midrule
Aloe-Beta-8B         & 8B  & Llama-3.1-8B-Instruct  & SFT + MERGE + DPO & 3 & chat & \yes   & \cite{garcia-gasullaetal2026aloe} \\
HuatuoGPT-o1-8B      & 8B  & Llama-3.1-8B-Instruct  & SFT + RL          & 2 & chat & \yes   & \cite{chenetal2024huatuogpto1} \\
Aloe-Beta-70B        & 70B & Llama-3.1-70B-Instruct & SFT + MERGE + DPO & 3 & chat & \yes   & \cite{garcia-gasullaetal2026aloe} \\
HuatuoGPT-o1-70B     & 70B & Llama-3.1-70B-Instruct & SFT + RL          & 2 & chat & \yes   & \cite{chenetal2024huatuogpto1} \\
MMed-Llama-3.1-70B   & 70B & Llama-3.1-70B          & CPT               & 1 & compl. & \plaus & \cite{qiuetal2024building} \\
MMed-Llama-3.1-70B-MCQA & 70B & Llama-3.1-70B       & CPT + SFT         & 2 & chat & \yes   & \cite{qiuetal2024building} \\
\midrule
\multicolumn{8}{@{}l}{\textit{Gemma-3 family}} \\
\midrule
MedGemma-4B-IT       & 4B  & Gemma-3-4B-PT  & CPT + SFT + RL & 3 & chat & \plaus & \cite{sellergrenetal2025medgemma} \\
MedGemma-27B-Text-IT & 27B & Gemma-3-27B-PT & CPT + SFT + RL & 3 & chat & \plaus & \cite{sellergrenetal2025medgemma} \\
\bottomrule
\end{tabular}%
}
\end{table*}

\subsection{Complete Metric Definitions}
\label{app:metric-definitions}

\subsubsection{Notation and Aligned Evaluation Populations}

For a domain-adapted/base pair, let $r\in\mathcal{R}=\{b,f\}$ index model role, where $b$ denotes the base model and $f$ its domain-adapted variant. Let $v$ index benchmark condition, with $v\in\mathcal{V}_{\mathrm{MA}}=\{\mathrm{NA},\mathrm{A},\mathrm{NAP},\mathrm{AP}\}$ for \MA\ and $v\in\mathcal{V}_{\mathrm{CD}}=\{\mathrm{Complete},\mathrm{Det},\mathrm{Undet}\}$ for \CD. Here, $\mathrm{Det}$ and $\mathrm{Undet}$ denote \CDID and \CDIU, respectively. Let $I_v$ be the aligned item set for condition $v$ and $n_v=|I_v|$. For item $i$, let $Y_{r,v,i}\in\{0,1\}$ indicate parsed-answer correctness and $Z_{r,v,i}\in\{0,1\}$ indicate abstention; for \CD, $Z_{r,v,i}=1$ indicates selection of ``Unable to determine.'' Let $P_{r,v,i,k}$ be the probability assigned to choice $k$, renormalized over the declared answer choices; $g_{v,i}$ the gold choice index; $\widehat{g}_{r,v,i}=\arg\max_kP_{r,v,i,k}$ the maximum-probability choice; and $C_{r,v,i}=\max_kP_{r,v,i,k}$ the associated confidence. For \MA\ perturbed conditions, let $\widetilde{Y}_{r,v,i}=\mathbf{1}[\widehat{g}_{r,v,i}=g^{\mathrm{retained}}_{v,i}]$ indicate agreement with the retained original key.

For a condition-specific metric $m_{r,v}$, every pair-level effect is defined as
\begin{equation}
\label{eq:app_delta}
    \Delta m_v=m_{f,v}-m_{b,v}.
\end{equation}
We suppress $v$ when the condition is clear or when $m$ combines multiple conditions. The absolute change is the primary estimand and is expressed in percentage points for proportion-valued metrics. Supplementary relative changes are
\begin{equation}
\label{eq:app_relative_delta}
    \Delta_{\mathrm{rel}}m_v=\frac{m_{f,v}-m_{b,v}}{m_{b,v}},
\end{equation}
where $m_{b,v}>0$. Relative changes are calculated within each model pair before aggregation. Pairs with zero or near-zero base values are omitted from the relative-change view rather than assigned unstable values.

Behavioral metrics use all aligned items, including parse failures. Probability metrics use the jointly captured population
\begin{equation}
\label{eq:joint_population}
\begin{aligned}
    I_{v,\mathrm{joint}}
    &=\{i\in I_v:\\
    &\quad\text{valid choice probabilities for both roles,}\\
    &\quad g_{v,i}\text{ is valid}\}.
\end{aligned}
\end{equation}
\MA\ uses five choices for \MANA\ and \MANAP\ and six choices for \MAA\ and \MAAP. \CD\ uses ``Met,'' ``Not Met,'' and ``Unable to determine.'' Primary \CD\ probability metrics use probabilities captured at the generated answer token and renormalized across these three labels.

\subsubsection{Accuracy and Retained-Key Agreement}

On a condition with a clinically valid gold answer, accuracy is
\begin{equation}
\label{eq:app_accuracy}
    \operatorname{Accuracy}_{r,v}
    =
    \frac{1}{n_v}
    \sum_{i\in I_v}Y_{r,v,i}.
\end{equation}
This definition is used for \MANA\ and \MAA\ and for all \CD\ conditions. On \MAA, selecting the abstention option is not counted as correct because the original question remains answerable.

\MANAP\ and \MAAP\ retain the original answer key after evidence has been removed. Agreement with this key is
\begin{equation}
\label{eq:app_retained_key}
    R_{r,v}
    =
    \frac{1}{n_v}
    \sum_{i\in I_v}
    \mathbf{1}
    \left[
        \widetilde{Y}_{r,v,i}=1
    \right],
    \qquad
    v\in\{\mathrm{NAP},\mathrm{AP}\}.
\end{equation}
Retained-key agreement measures whether a model preserves the original answer following evidence removal. It is a perturbation-response diagnostic and is not interpreted as clinical accuracy on \MANAP\ or \MAAP.

\subsubsection{Abstention, Coverage, and Selective Accuracy}

The abstention or ``Unable to determine'' rate is
\begin{equation}
\label{eq:app_abstention}
    U_{r,v}
    =
    \frac{1}{n_v}
    \sum_{i\in I_v}Z_{r,v,i}.
\end{equation}
Coverage is the fraction of questions receiving a non-abstaining answer:
\begin{equation}
\label{eq:app_coverage}
    \operatorname{Coverage}_{r,v}
    =1-U_{r,v}
    =
    \frac{1}{n_v}
    \sum_{i\in I_v}(1-Z_{r,v,i}).
\end{equation}
Selective accuracy conditions on the questions a model answered:
\begin{equation}
\label{eq:app_selective_accuracy}
    \operatorname{SelAcc}_{r,v}
    =
    \frac{
        \sum_{i\in I_v}Y_{r,v,i}(1-Z_{r,v,i})
    }{
        \sum_{i\in I_v}(1-Z_{r,v,i})
    }.
\end{equation}
The denominator can differ between the base and domain-adapted models. Selective accuracy is undefined if a model answers no questions in the condition.

\subsubsection{Behavioral Overconfidence and Underconfidence}

Behavioral overconfidence means answering when the benchmark indicates that the evidence is insufficient; behavioral underconfidence means abstaining when the benchmark indicates that an answer or determination remains possible. These are decision-level error rates and are distinct from probability-based confidence and calibration.

For \MA,
\begin{equation}
\label{eq:app_ma_overconfidence}
    \mathrm{OC}^{\mathrm{MA}}_r
    =
    \Pr(Z_{r,\mathrm{AP},i}=0\mid i\in I_{\mathrm{AP}})
    =1-U_{r,\mathrm{AP}},
\end{equation}
and
\begin{equation}
\label{eq:app_ma_underconfidence}
    \mathrm{UC}^{\mathrm{MA}}_r
    =
    \Pr(Z_{r,\mathrm{A},i}=1\mid i\in I_{\mathrm{A}})
    =U_{r,\mathrm{A}}.
\end{equation}
Thus, \MA\ behavioral overconfidence is answering a \MAAP\ question, while behavioral underconfidence is abstaining on an answerable \MAA\ question. {Because benchmark annotation indicates that 11 of 49 labeled \MAAP\ questions may remain answerable after perturbation, $\mathrm{OC}^{\mathrm{MA}}$ is a benchmark-defined behavioral error rate rather than a definitive clinical judgment for every item.}

For \CD,
\begin{equation}
\label{eq:app_cd_overconfidence}
\begin{aligned}
    \mathrm{OC}^{\mathrm{CD}}_r
    &=\Pr(Z_{r,\mathrm{Undet},i}=0\\
    &\qquad\mid i\in I_{\mathrm{Undet}})
      =1-U_{r,\mathrm{Undet}},
\end{aligned}
\end{equation}
and
\begin{equation}
\label{eq:app_cd_underconfidence}
    \mathrm{UC}^{\mathrm{CD}}_r
    =
    \Pr(Z_{r,\mathrm{Det},i}=1\mid i\in I_{\mathrm{Det}})
    =U_{r,\mathrm{Det}}.
\end{equation}
\CD\ behavioral overconfidence means answering a \CDIU\ case with ``Met'' or ``Not Met''. Behavioral underconfidence is selecting ``Unable to determine'' on a \CDID case.

For either benchmark, positive $\Delta\mathrm{OC}$ or $\Delta\mathrm{UC}$ represents movement in an undesirable direction. Because behavioral overconfidence is the exact complement of appropriate abstention, testing $\Delta\mathrm{OC}=0$ is equivalent to testing the corresponding appropriate-abstention change. Behavioral overconfidence is therefore not entered as an additional independent claim in the primary correction family.

\subsubsection{Informedness}

As defined in the main text, informedness $J$ is given for a binary diagnostic test by
\begin{equation}
\label{eq:app_informedness_j}
    J=\operatorname{Sensitivity}+\operatorname{Specificity}-1.
\end{equation}
For our application, an abstention-appropriate condition is treated as the positive class and abstention as the positive decision. The resulting \MA\ statistic is
\begin{equation}
\label{eq:app_ma_j}
    J_r^{\mathrm{MA}}
    =U_{r,\mathrm{AP}}-U_{r,\mathrm{A}}
    =1-\mathrm{OC}^{\mathrm{MA}}_r-\mathrm{UC}^{\mathrm{MA}}_r.
\end{equation}
The corresponding \CD\ statistic is
\begin{equation}
\label{eq:app_cd_j}
    J_r^{\mathrm{CD}}
    =U_{r,\mathrm{Undet}}-U_{r,\mathrm{Det}}
    =1-\mathrm{OC}^{\mathrm{CD}}_r-\mathrm{UC}^{\mathrm{CD}}_r.
\end{equation}
Larger $J$ indicates greater informedness. A negative $\Delta J=J_f-J_b$ indicates that domain adaptation reduced informedness.

\subsubsection{\MA\ Responsiveness and Targeting}

The retained-key drop following evidence removal is
\begin{equation}
\label{eq:app_key_drop}
    D_r=\operatorname{Accuracy}_{r,\mathrm{NA}}-R_{r,\mathrm{NAP}},
\end{equation}
and the corresponding abstention response is
\begin{equation}
\label{eq:app_abstention_response}
    G_r=U_{r,\mathrm{AP}}-U_{r,\mathrm{A}}.
\end{equation}
The responsiveness ratio is
\begin{equation}
\label{eq:app_responsiveness_ratio}
    \mathrm{RR}_r=\frac{G_r}{D_r},\qquad D_r>0.
\end{equation}
It is undefined when $D_r\leq0$. A value of one indicates that the increase in abstention equals the reduction in retained-key agreement; values below one indicate behavioral under-response to the measured effect of evidence removal.

To assess whether \MAAP\ abstention is targeted to the questions on which perturbation changes the model's retained answer, define
\begin{equation}
\label{eq:app_targeting_sets}
\begin{aligned}
    \mathcal{L}_r
    &=\{i:Y_{r,\mathrm{NA},i}=1,\;\widetilde{Y}_{r,\mathrm{NAP},i}=0\},\\
    \mathcal{H}_r
    &=\{i:Y_{r,\mathrm{NA},i}=1,\;\widetilde{Y}_{r,\mathrm{NAP},i}=1\}.
\end{aligned}
\end{equation}
The targeting statistic is
\begin{equation}
\label{eq:app_targeting}
\begin{aligned}
    T_r
    &=\Pr(Z_{r,\mathrm{AP},i}=1\mid i\in\mathcal{L}_r)\\
    &\quad-\Pr(Z_{r,\mathrm{AP},i}=1\mid i\in\mathcal{H}_r).
\end{aligned}
\end{equation}
Positive targeting indicates that the model abstains more often when evidence removal changes its retained answer.

\subsubsection{Coverage-Transition Partitions}

On \MAA, items are partitioned according to base and domain-adapted model abstention:
\begin{equation}
\label{eq:app_transition_sets}
\begin{aligned}
    \mathcal{N}
    &=\{i:Z_{b,\mathrm{A},i}=1,\;Z_{f,\mathrm{A},i}=0\},\\
    \mathcal{B}
    &=\{i:Z_{b,\mathrm{A},i}=0,\;Z_{f,\mathrm{A},i}=0\},\\
    \mathcal{M}
    &=\{i:Z_{b,\mathrm{A},i}=0,\;Z_{f,\mathrm{A},i}=1\},\\
    \mathcal{Z}
    &=\{i:Z_{b,\mathrm{A},i}=1,\;Z_{f,\mathrm{A},i}=1\}.
\end{aligned}
\end{equation}
Here $\mathcal{N}$, $\mathcal{B}$, $\mathcal{M}$, and $\mathcal{Z}$ denote newly answered, always answered, newly abstained, and both abstained questions, respectively. These sets are mutually exclusive and exhaustive over aligned \MAA\ questions.

For newly answered questions, only the domain-adapted model produces an answer. We calculate
\begin{equation}
\label{eq:app_newly_answered}
\begin{aligned}
    \operatorname{AccNew}_f
    &=\frac{1}{|\mathcal{N}|}\sum_{i\in\mathcal{N}}Y_{f,\mathrm{A},i},\\
    \operatorname{AccAlways}_f
    &=\frac{1}{|\mathcal{B}|}\sum_{i\in\mathcal{B}}Y_{f,\mathrm{A},i}.
\end{aligned}
\end{equation}
with gap
\begin{equation}
\label{eq:app_newly_answered_gap}
    \operatorname{Gap}_{\mathcal{N}}
    =\operatorname{AccNew}_f-\operatorname{AccAlways}_f.
\end{equation}

For newly abstained questions, only the base model produces an answer. We calculate
\begin{equation}
\label{eq:app_newly_abstained}
\begin{aligned}
    \operatorname{AccNewAbstained}_b
    &=\frac{1}{|\mathcal{M}|}\sum_{i\in\mathcal{M}}Y_{b,\mathrm{A},i},\\
    \operatorname{AccAlways}_b
    &=\frac{1}{|\mathcal{B}|}\sum_{i\in\mathcal{B}}Y_{b,\mathrm{A},i}.
\end{aligned}
\end{equation}
with gap
\begin{equation}
\label{eq:app_newly_abstained_gap}
    \operatorname{Gap}_{\mathcal{M}}
    =\operatorname{AccNewAbstained}_b-\operatorname{AccAlways}_b.
\end{equation}
Negative values indicate that the coverage-changing partition is less accurate than the always-answered partition.

The partitions defined on \MAA\ are transferred to \MANA, where neither model can abstain. For $\mathcal{P}\in\{\mathcal{B},\mathcal{N},\mathcal{M},\mathcal{Z}\}$,
\begin{equation}
\label{eq:app_na_partition_accuracy}
\begin{aligned}
    \operatorname{Accuracy}^{\mathrm{NA}}_{r,\mathcal{P}}
    &=\frac{1}{|\mathcal{P}|}
      \sum_{i\in\mathcal{P}}Y_{r,\mathrm{NA},i},\\
    \Delta_{\mathcal{P}}
    &=\operatorname{Accuracy}^{\mathrm{NA}}_{f,\mathcal{P}}
      -\operatorname{Accuracy}^{\mathrm{NA}}_{b,\mathcal{P}}.
\end{aligned}
\end{equation}
Comparing accuracy across partitions asks whether coverage-changing questions are intrinsically harder; comparing the base and domain-adapted roles within a partition asks whether domain adaptation changes capability on those same questions.

\subsubsection{Probability, Calibration, and Predictive Sharpness Metrics}

All probability metrics are calculated on $I_{v,\mathrm{joint}}$; let $n=|I_{v,\mathrm{joint}}|$. Mean maximum-choice confidence is
\begin{equation}
\label{eq:app_mean_confidence}
\begin{aligned}
    \operatorname{Confidence}_{r,v}
    &=\frac{1}{n}\sum_{i\in I_{v,\mathrm{joint}}}C_{r,v,i},\\
    C_{r,v,i}&=\max_kP_{r,v,i,k}.
\end{aligned}
\end{equation}
Higher confidence can accompany either increased capability or more confident errors and is not, by itself, evidence of better calibration.

For $K_v$ choices, per-item normalized entropy and its mean are
\begin{equation}
\label{eq:app_entropy}
\begin{aligned}
    H_{r,v,i}
    &=-\frac{\sum_{k=1}^{K_v}P_{r,v,i,k}\log P_{r,v,i,k}}{\log K_v},\\
    \overline{H}_{r,v}
    &=\frac{1}{n}\sum_{i\in I_{v,\mathrm{joint}}}H_{r,v,i}.
\end{aligned}
\end{equation}
where $0\log0=0$. Normalized entropy lies in $[0,1]$; lower entropy indicates a sharper predictive distribution, not necessarily better calibration.

Expected calibration error is calculated using ten equal-width confidence bins. Let $\mathcal{C}_j$ contain the jointly captured items whose maximum-choice confidence falls in bin $j$. Then
\begin{equation}
\label{eq:app_ece}
\begin{aligned}
    a_{r,v,j}
    &=\frac{1}{|\mathcal{C}_j|}
      \sum_{i\in\mathcal{C}_j}\mathbf{1}[\widehat{g}_{r,v,i}=g_{v,i}],\\
    c_{r,v,j}
    &=\frac{1}{|\mathcal{C}_j|}
      \sum_{i\in\mathcal{C}_j}C_{r,v,i},\\
    \operatorname{ECE10}_{r,v}
    &=\sum_{j=1}^{10}\frac{|\mathcal{C}_j|}{n}
      |a_{r,v,j}-c_{r,v,j}|.
\end{aligned}
\end{equation}
Empty bins contribute zero. Lower ECE10 indicates closer agreement between probability-based confidence and empirical correctness.

\begin{figure*}[htbp]
    \centering
    \includegraphics[width=\textwidth]{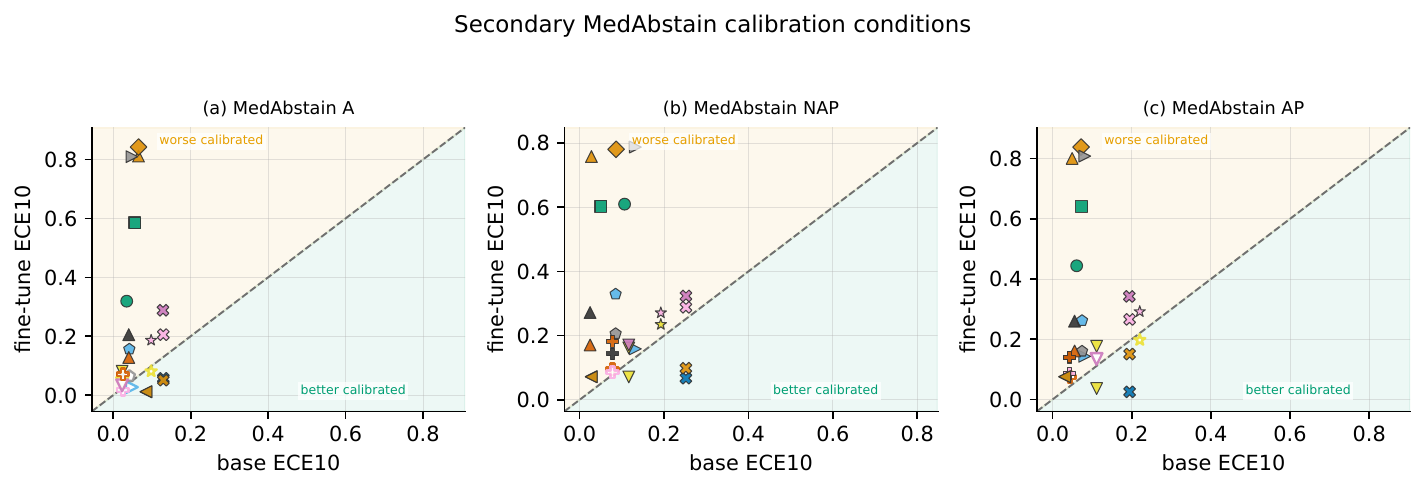}
    \caption{Fine-tune versus base ECE10 for each model pair on the secondary conditions \MAA, \MANAP, and \MAAP. Color encodes fine-tuning recipe, marker shape encodes base-model family, and the diagonal marks unchanged calibration.}
    \label{fig:ece_identity_appendix}
\end{figure*}

Confidence-based discrimination is measured using
\begin{equation}
\label{eq:app_auroc}
\begin{aligned}
    \operatorname{AUROC}_{r,v}
    &=\Pr(C_{rv i^+}>C_{rv i^-})\\
    &\quad+\frac{1}{2}\Pr(C_{rv i^+}=C_{rv i^-}).
\end{aligned}
\end{equation}
where $i^+$ and $i^-$ are sampled from correct and incorrect maximum-probability predictions, respectively. AUROC is undefined when all included predictions have the same correctness label.

Let
\begin{equation}
\label{eq:app_shared_wrong_set}
\begin{aligned}
    \mathcal{E}_{bf,v}
    =\{i\in I_{v,\mathrm{joint}}:\;&
      \widehat{g}_{b,v,i}\neq g_{v,i},\\
      &\widehat{g}_{f,v,i}\neq g_{v,i}\}.
\end{aligned}
\end{equation}
Shared-wrong confidence is
\begin{equation}
\label{eq:app_shared_wrong_confidence}
    W^{\mathrm{shared}}_{r,v}
    =
    \frac{1}{|\mathcal{E}_{bf,v}|}
    \sum_{i\in\mathcal{E}_{bf,v}}C_{r,v,i}.
\end{equation}
Both roles are evaluated on exactly the same incorrect questions. Consequently, $\Delta W^{\mathrm{shared}}>0$ indicates increased confidence on errors independently of changes in overall accuracy.

For completeness, supplementary probability outcomes include the full-vector multiclass Brier score
\begin{equation}
\label{eq:app_brier}
    \operatorname{Brier}_{r,v}
    =
    \frac{1}{n}
    \sum_{i\in I_{v,\mathrm{joint}}}
    \sum_{k=1}^{K_v}
    \left(P_{r,v,i,k}-\mathbf{1}[k=g_{v,i}]\right)^2,
\end{equation}
and negative log likelihood
\begin{equation}
\label{eq:app_nll}
    \operatorname{NLL}_{r,v}
    =
    -\frac{1}{n}
    \sum_{i\in I_{v,\mathrm{joint}}}
    \log\left(\max(P_{r,v,i,g_{v,i}},10^{-6})\right).
\end{equation}
Lower Brier score and negative log likelihood indicate better calibration.

\subsection{Complete Statistical Methods}
\label{app:statistical-methods}

\subsubsection{Units of Inference}

We use two distinct units of inference. Within a domain-adapted/base pair, the benchmark item is the unit of analysis. These tests determine whether one specific pair differs on aligned benchmark questions. Across the cohort, the domain-adapted/base pair is the unit of inference. Each pair contributes one $\Delta m_j$, and cohort-level claims evaluate the distribution of these pair-level effects. Pooling benchmark questions across model pairs would treat correlated observations generated by the same models as independent evidence and would constitute pseudo-replication.

\subsubsection{Exact Within-Pair Tests}

For aligned binary outcomes, including abstention, ``Unable to determine,'' correctness, and forced-answer partition accuracy, we use the exact McNemar test. Let $n_{10}$ denote the number of aligned questions with a positive base outcome and negative fine-tuned outcome, and $n_{01}$ the reverse transition. Under the null hypothesis of equal marginal probabilities,
\begin{equation}
\label{eq:app_mcnemar_null}
    X\mid(n_{10}+n_{01})
    \sim
    \operatorname{Binomial}(n_{10}+n_{01},0.5).
\end{equation}
The two-sided exact p-value is
\begin{equation}
\label{eq:app_mcnemar_p}
\begin{aligned}
    p_{\mathrm{McNemar}}
    =\min\bigl[1,2\min\{&
        \Pr[X\leq\min(n_{10},n_{01})],\\
        &\Pr[X\geq\max(n_{10},n_{01})]\}\bigr].
\end{aligned}
\end{equation}
Only discordant outcomes contribute to this test.

The newly answered and newly abstained gaps compare correctness across disjoint item partitions. We use a two-sided Fisher exact test on the corresponding correct/incorrect $2\times2$ table. If $N$ is the total number of observations, $K$ the total number correct, $n_1$ the size of the first partition, and $x$ its number correct, then under fixed margins
\begin{equation}
\label{eq:app_fisher}
    \Pr(X=x)
    =
    \frac{
        \binom{K}{x}\binom{N-K}{n_1-x}
    }{
        \binom{N}{n_1}
    }.
\end{equation}
The two-sided p-value sums the probabilities of tables with the same margins that are at least as extreme as the observed table.

\subsubsection{Paired Item Bootstrap}

For continuous metrics and cross-condition quantities, we use a paired item bootstrap. Aligned item indices are sampled with replacement, and the base and fine-tuned roles are evaluated using the same sampled indices. Cross-condition \MA\ statistics use the same sampled underlying item indices across \MANA, \MAA, \MANAP, and \MAAP\ to preserve item correspondence.

We use $N_{\mathrm{boot}}=1{,}000$ replications for most metrics and $N_{\mathrm{boot}}=500$ for ECE10 and AUROC. Let $\Delta m^*$ denote the effect recomputed in one bootstrap replication and $Q_a(\Delta m^*)$ the empirical $a$-quantile of the bootstrap effects. The percentile confidence interval is
\begin{equation}
\label{eq:app_bootstrap_ci}
    \operatorname{CI}_{95\%}
    =
    \left[
        Q_{0.025}(\Delta m^*),
        Q_{0.975}(\Delta m^*)
    \right].
\end{equation}
The associated two-sided percentile-sign p-value is
\begin{equation}
\label{eq:app_bootstrap_p}
    p_{\mathrm{boot}}
    =
    2\min\left\{
        \Pr(\Delta m^*\leq0),
        \Pr(\Delta m^*\geq0)
    \right\}.
\end{equation}
This is a percentile-sign p-value obtained from the bootstrap distribution rather than a permutation p-value. Its minimum possible non-zero value is $2/N_{\mathrm{boot}}$.

If a bootstrap replication produces an undefined statistic, such as a responsiveness ratio with $D_r\leq0$ or an empty conditional partition, that replication is excluded. The interval is reported as undefined if fewer than 5\% of the requested replications produce a valid estimate.

\subsubsection{Across-Pair Tests}

For each outcome and condition, model pair $j\in\{1,\ldots,N_{\mathrm{pair}}\}$ contributes a single effect $\Delta m_{v,j}$. We suppress $v$ when the condition is clear. We use a two-sided Wilcoxon signed-rank test to evaluate whether the distribution of pair-level effects is symmetric around zero. Let $R_j$ be the rank of $|\Delta m_{v,j}|$ among the non-zero effects. The positive-rank statistic is
\begin{equation}
\label{eq:app_wilcoxon}
    W^+
    =
    \sum_{j:\Delta m_{v,j}>0}R_j.
\end{equation}
We report the median pair-level effect as the primary cohort effect-size summary.

Associations between two continuous or ordinal pair-level variables are tested using permutation inference. Let $s_{\mathrm{obs}}$ denote the observed statistic, $s_h^*$ the statistic after the $h$th label permutation, and $N_{\mathrm{perm}}=2{,}000$ the number of permutations. The two-sided permutation p-value is
\begin{equation}
\label{eq:app_permutation}
    p_{\mathrm{perm}}
    =
    \frac{
        1+
        \sum_{h=1}^{N_{\mathrm{perm}}}
        \mathbf{1}
        \left[|s_h^*|\geq|s_{\mathrm{obs}}|\right]
    }{
        N_{\mathrm{perm}}+1
    }.
\end{equation}
The Pearson correlation coefficient $r_{\mathrm{P}}$ is used for the primary relationship between capability change and caution change. Spearman correlation $\rho$ or Kendall correlation $\tau$ is used for ordinal recipe features and hyperparameters. Permutation group contrasts are used for exploratory recipe-family comparisons.

\subsubsection{False-Discovery-Rate Correction}

We control the false-discovery rate using the Benjamini--Hochberg (BH) procedure. Within a declared family of $M$ tests, let $p_i$ denote the unadjusted two-sided p-value for test $i$, and let $p_{(1)}\leq\cdots\leq p_{(M)}$ denote those p-values ordered from smallest to largest. For the test occupying rank $i$, $q_{\mathrm{BH},(i)}$ denotes its BH-adjusted q-value:
\begin{equation}
\label{eq:app_bh}
    q_{\mathrm{BH},(i)}
    =
    \min_{\ell\geq i}
    \left\{
        \frac{M}{\ell}p_{(\ell)}
    \right\},
\end{equation}
where $\ell$ indexes ordered positions at or above rank $i$. Adjusted values are constrained to be monotonic and no greater than one. In the main text, $q_{\mathrm{BH}}$ denotes the adjusted q-value for the specific reported test; the rank subscript is omitted for readability.

The correction families are:
\begin{itemize}
    \item \textbf{Within-pair metrics:} the available fine-tune/base pairs within one metric column;
    \item \textbf{Primary hypotheses:} 15 cohort-level claims across appropriate-caution erosion (\textbf{H1}), informedness change (\textbf{H2}), coverage-change accuracy (\textbf{H3}), commitment and calibration (\textbf{H4}), and capability--caution coupling and heterogeneity (\textbf{H5});
    \item \textbf{Commitment and calibration hypothesis (H4) family:} five calibration and sharpness metrics on \MANA\ and five on \CDC;
    \item \textbf{Moderator analyses:} separately declared recipe, alignment, and size families.
\end{itemize}

The 15-claim primary family contains two appropriate-caution erosion (\textbf{H1}) outcomes, two behavioral-underconfidence outcomes, two informedness-change (\textbf{H2}) outcomes, two coverage-change accuracy (\textbf{H3}) outcomes, five \MANA\ commitment and calibration (\textbf{H4}) outcomes, and two capability--caution coupling and heterogeneity (\textbf{H5}) outcomes. Behavioral overconfidence is not entered separately because it is the exact complement of the corresponding appropriate-abstention outcome and would duplicate the same test.

\subsubsection{Dependence among Domain-Adapted/Base Pairs}

The 25 domain-adapted models descend from 11 base checkpoints, with several bases anchoring multiple adapted variants. A Wilcoxon test treats the 25 pair-level effects as independent and may overstate precision when adapted models sharing a base checkpoint also share base-specific behavior.

We therefore accompany primary cohort-level results with a hierarchical base-cluster bootstrap. Each replication first samples base checkpoints with replacement, then samples adapted descendants with replacement within each selected base, and finally recomputes the cohort statistic. This procedure preserves dependence within shared-base clusters and prevents a base checkpoint with many descendants from dominating each replication. We report the resulting p-value as $p_{\mathrm{cluster}}$.

A primary result is considered robust when its direction is consistent, $q_{\mathrm{BH}}<0.05$, and $p_{\mathrm{cluster}}<0.05$. Subset robustness analyses are interpreted alongside their effect directions and sample sizes.

\subsubsection{Heterogeneity}
\label{app:results-hetero}

We test whether pair-level effects are compatible with a common effect using Cochran's $Q$. {For each pair, the analysis approximates $\mathrm{SE}_j=(\mathrm{CI}^{\mathrm{hi}}_j-\mathrm{CI}^{\mathrm{lo}}_j)/(2\times1.96)$ from the paired-item percentile-bootstrap interval and sets $w_j=1/\mathrm{SE}_j^2$; pairs with non-finite or non-positive standard errors are omitted. This normal-width approximation can be inaccurate for asymmetric or boundary-clipped percentile intervals, so $Q$ and $I^2$ are descriptive evidence of heterogeneity rather than precise variance-component estimates.} For $N_{\mathrm{pair}}$ pair-level estimates $\Delta_j$ with inverse-variance weights $w_j$, the weighted mean is
\begin{equation}
\label{eq:app_weighted_mean}
    \overline{\Delta}_w
    =
    \frac{
        \sum_{j=1}^{N_{\mathrm{pair}}}w_j\Delta_j
    }{
        \sum_{j=1}^{N_{\mathrm{pair}}}w_j
    }.
\end{equation}
Cochran's statistic is
\begin{equation}
\label{eq:app_cochran_q}
    Q
    =
    \sum_{j=1}^{N_{\mathrm{pair}}}
    w_j(\Delta_j-\overline{\Delta}_w)^2.
\end{equation}
Under the homogeneous-effect null hypothesis, $Q$ is compared with a $\chi^2$ distribution with $N_{\mathrm{pair}}-1$ degrees of freedom. The proportion of total variation attributable to between-pair heterogeneity is
\begin{equation}
\label{eq:app_i_squared}
    I^2
    =
    \max\left\{
        0,
        \frac{Q-(N_{\mathrm{pair}}-1)}{Q}
    \right\}
    \times100\%.
\end{equation}
Large $I^2$ values indicate that variation across fine-tune/base pairs is not adequately described by a single common effect. For \MAAP, $Q=642.92$ on 24 degrees of freedom ($p=2.4\times10^{-120}$) and $I^2=96.27\%$. For \CDIU, $Q=363.09$ on 24 degrees of freedom ($p=2.7\times10^{-62}$) and $I^2=93.39\%$.

\subsubsection{Robustness and Sensitivity Analyses}
\label{app:robustness-sensitivity}

We perform the following sensitivity analyses where applicable:
\begin{itemize}
    \item \textbf{Confirmed-overlap-excluded:} exclude pairs with confirmed benchmark overlap while retaining pairs whose overlap is only plausible;
    \item \textbf{Thin-cell-excluded:} exclude transition analyses with fewer than 15 questions in the relevant newly answered or newly abstained partition; and
    \item \textbf{Relative-change analysis:} repeat summaries using $(m_{f,v}-m_{b,v})/m_{b,v}$ where the base value is sufficiently above zero.
\end{itemize}
Quality flags are displayed and evaluated through sensitivity analyses.

Excluding the seven pairs with confirmed MedQA overlap leaves 18 pairs. Appropriate abstention decreases in 14 of those 18 pairs, with a median change of −3.92 pp (Wilcoxon p=0.0346; base-clustered p=0.046; 95\% CI −6.36 to −0.70 pp). Only two pairs have confirmed no overlap with MedQA, so this sensitivity excludes only those models where overlap is confirmed rather than all models where overlap is possible.

Excluding four merged descendants and Palmyra (due to inferred lineage) leaves 20 pairs over 11 base checkpoints. In this case, appropriate abstention decreases in 17 of 20 pairs, with a median change of −3.43 pp (Wilcoxon p=0.0172; base-clustered p=0.003; 95\% CI −5.66 to −0.89 pp).

At each base checkpoint's observed coverage, the median selective-risk difference (adapted minus base) is −1.42 pp (Wilcoxon p=0.287; base-clustered p=0.264; 95\% CI −6.48 to +2.91 pp). The analysis does not show a generally worse selective ranking policy after adaptation; it instead shows an operating-point shift into answering harder items.

\end{document}